\documentclass[10pt,twocolumn,letterpaper]{article}

\usepackage[pagenumbers]{cvpr} %

\usepackage{bm}
\usepackage{multirow}
\usepackage[dvipsnames]{xcolor}

\definecolor{cvprblue}{rgb}{0.21,0.49,0.74}
\usepackage[pagebackref,breaklinks,colorlinks,citecolor=cvprblue]{hyperref}

\title{\model: Semantic Edits on 3D Image GANs}

\author{\stepcounter{footnote}\vspace{1mm}Enis Simsar\thanks{Conducted this research as part of studies at TUM.} $^{,1,2}$
\hspace{0.75cm}
Alessio Tonioni$^{3}$
\hspace{0.75cm}
Evin Pınar Örnek$^{2}$
\hspace{0.75cm}
Federico Tombari$^{2,3}$
\\
$^1$ETH Zürich - DALAB
\hspace{2.5em} 
$^2$Technical University of Munich
\hspace{2.5em} 
$^3$Google Switzerland
}

\begin{document}
\maketitle

\begin{abstract}
    3D GANs have the ability to generate latent codes for entire 3D volumes rather than only 2D images. These models offer desirable features like high-quality geometry and multi-view consistency, but, unlike their 2D counterparts, complex semantic image editing tasks for 3D GANs have only been partially explored.
    To address this problem, we propose \model{}, a semantic edit approach based on latent space discovery that can be used with any off-the-shelf 3D or 2D GAN model and on any dataset.
    \model{} relies on identifying the latent code dimensions corresponding to specific attributes by feature ranking using a random forest classifier. It then performs the edit by swapping the selected dimensions of the image being edited with the ones from an automatically selected reference image. %
    Compared to other latent space control-based edit methods, which were mainly designed for 2D GANs, our method on 3D GANs provides remarkably consistent semantic edits in a disentangled manner and outperforms others both qualitatively and quantitatively.
    We show results on seven 3D GANs (\pigan{}, GIRAFFE, StyleSDF, MVCGAN, EG3D, StyleNeRF, and VolumeGAN) and on five datasets (FFHQ, AFHQ, Cats, MetFaces, and CompCars). 
\end{abstract}

\section{Introduction}
\label{sec:introduction}
\begin{figure}[ht!]
\centering
\small

\begin{minipage}{.23\linewidth}
    \centering
    \textbf{Input}
\end{minipage}
\begin{minipage}{.23\linewidth}
    \centering
    \textbf{View \#1}
\end{minipage}
\begin{minipage}{.23\linewidth}
    \centering
    \textbf{View \#2}
\end{minipage}
\begin{minipage}{.23\linewidth}
    \centering
    \textbf{View \#3}
\end{minipage}
\begin{minipage}{.04\linewidth}
    \raggedright
    \rotatebox[origin=lc]{270}{\centering \phantom{\footnotesize{\textbf{g}}}}
\end{minipage}

\vspace{0.1cm}

\begin{minipage}{.23\linewidth}
    \centering
    \includegraphics[width=\linewidth]{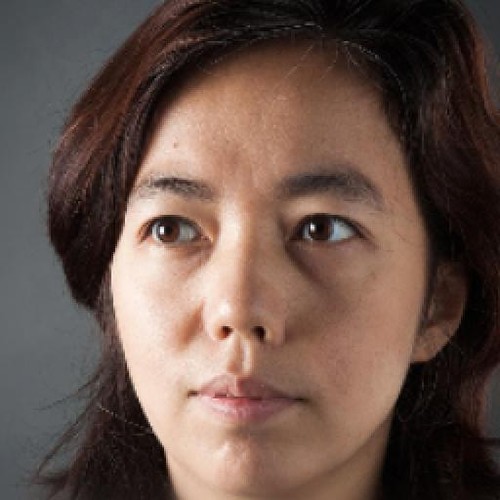}
\end{minipage}
\begin{minipage}{.23\linewidth}
    \centering
    \includegraphics[width=\linewidth]{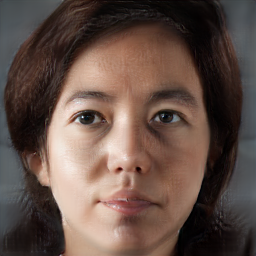}
\end{minipage}
\begin{minipage}{.23\linewidth}
    \centering
    \includegraphics[width=\linewidth]{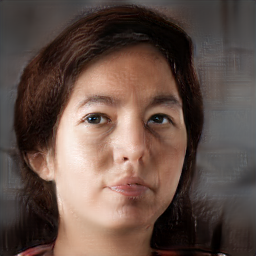}
\end{minipage}
\begin{minipage}{.23\linewidth}
    \centering
    \includegraphics[width=\linewidth]{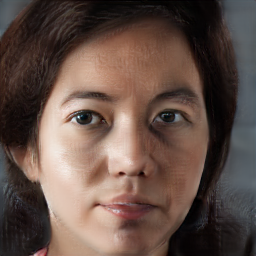}
\end{minipage}
\begin{minipage}{.04\linewidth}
    \raggedright
    \rotatebox[origin=lc]{270}{\centering \phantom{\footnotesize{\textbf{g}}}}
\end{minipage}

\begin{minipage}{.23\linewidth}
    \centering
    \textbf{StyleFlow\\\cite{abdal2021styleflow}}
\end{minipage}
\begin{minipage}{.23\linewidth}
    \centering
    \includegraphics[width=\linewidth]{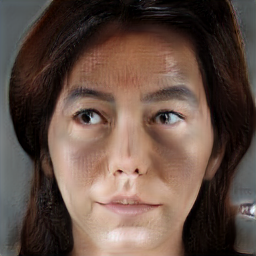}
\end{minipage}
\begin{minipage}{.23\linewidth}
    \centering
    \includegraphics[width=\linewidth]{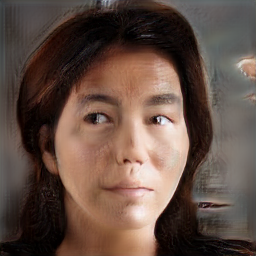}
\end{minipage}
\begin{minipage}{.23\linewidth}
    \centering
    \includegraphics[width=\linewidth]{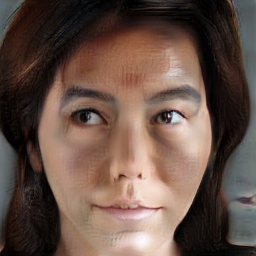}
\end{minipage}
\begin{minipage}{.04\linewidth}
    \raggedright
    \rotatebox[origin=lc]{270}{\centering \footnotesize{\textbf{Smiling (+)}}}
\end{minipage}

\begin{minipage}{.23\linewidth}
    \centering
    \textbf{Ours}
\end{minipage}
\begin{minipage}{.23\linewidth}
    \centering
    \includegraphics[width=\linewidth]{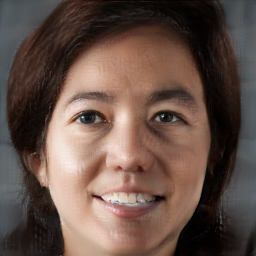}
\end{minipage}
\begin{minipage}{.23\linewidth}
    \centering
    \includegraphics[width=\linewidth]{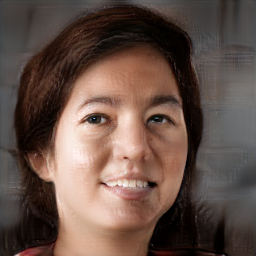}
\end{minipage}
\begin{minipage}{.23\linewidth}
    \centering
    \includegraphics[width=\linewidth]{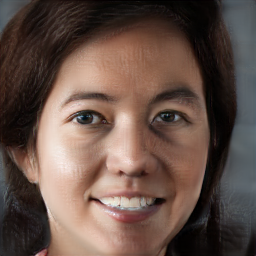}
\end{minipage}
\begin{minipage}{.04\linewidth}
    \raggedright
    \rotatebox[origin=lc]{270}{\centering \footnotesize{\textbf{Smiling (+)}}}
\end{minipage}

\caption{Our method (LatentSwap3D) inverts a given image in the latent space of a pre-trained MVCGAN~\cite{zhang2022multi} on FFHQ~\cite{karras2019style} while enabling novel view synthesis. 
Rows two and three show a comparison on attribute editing (\eg{,} smiling) between StyleFlow~\cite{abdal2021styleflow} and ours on a real face.\label{fig:teaser}}

\end{figure}

3D Generative Adversarial Networks~(3D GANs) have broad applications in fields like computer graphics and augmented and virtual reality~(AR/VR) thanks to their ability to synthesize photorealistic images with explicit camera pose control. 3D GANs could provide greater control over the subject to be edited by ensuring multi-view consistency when combined with semantic attribute editing. Such capabilities empower various applications ranging from realistic virtual try-on, and virtual product placement in movies or video games, to architectural design. For instance, they can enable \textit{changing hair color}, \textit{wearing eyeglasses}, and \textit{smiling} in the case of face generation or \textit{changing fur color and/or breed type} in the context of animal generation. 

Existing image editing methods have primarily focused on 2D GANs, and they provide robust control over attributes by manipulating latent spaces~\cite{jahanian2019steerability,shen2021closed,shen2020interfacegan,patashnik2021styleclip,yuksel2021latentclr}. However, current editing methods for 3D GANs are: limited to editing pose, expression, and illumination~\cite{liu20223dfmgan,mostgan,stylerig,tang2022explicitly,deng2020disentangled,tewari2020pie,sun2022controllable}, require training of the generator from scratch~\cite{deng2020disentangled,KowalskiECCV2020,shoshan2021gan_control,mostgan,sun2022fenerf,sun2022ide,chen2022sem2nerf,jiang2022nerffaceediting,deng2023pix2pix3d,tang2022explicitly} or require additional semantic segmentation maps as conditioning~\cite{sun2022fenerf,sun2022ide,chen2022sem2nerf,jiang2022nerffaceediting,deng2023pix2pix3d}. Therefore, exploring and controlling semantic attributes on latent spaces of any \emph{pre-trained} 3D GANs for attribute editing \emph{without the need to re-train or fine-tune the generator} remains an open research question. Although 2D editing methods may be effective for certain 3D GANs that inherit the latent space of StyleGAN, noticeably EG3D\cite{chan2021efficient}, they often lead to undesirable artifacts for other 3D GANs as shown in Fig.~\ref{fig:teaser}.
We argue that semantic attribute editing should perform as effectively on any 3D GAN model 
even if it does not inherit
StyleGAN latent space 
(\eg{,} GRAF, GIRAFFE, \pigan{}, MVCGAN, StyleSDF, and VolumeGAN)~\cite{graf,Niemeyer2020GIRAFFE,pigan,zhang2022multi,xu2021volumegan,or2021stylesdf}.

This work proposes a method to achieve multi-view consistent attribute editing on \emph{any} pre-trained 3D GAN, \ie{,} whether or not inheriting StyleGAN-based latent spaces. Our approach first explores 3D GANs' latent spaces. Then, it identifies latent dimensions that strongly correlate with the desired attribute. Finally, it performs edits by swapping the identified codes with the corresponding codes from the automatically selected reference subject already possessing the desired attribute. Unlike linear operations or predicting the edited latent codes, our proposed swapping method ensures that the edited latent codes remain within the range of valid values expected by the generative model.
 
Like their 2D counterparts, 3D GANs expose various latent spaces that control image generation. 
Therefore, as a preliminary step to enable attribute edits, we find the most suitable latent space by measuring disentanglement, completeness, and informativeness~(DCI) metrics, as proposed in~\cite{eastwood2018framework} and firstly used in ~\cite{wu2021stylespace} to assess the quality of latent spaces of generative models. 
To identify which dimensions in the latent space control the presence or absence of a specific attribute, we employ a method that involves training a random forest~\cite{breiman2001random} with latent codes to perform regression for the presence or absence of the desired attribute.
The learned random forest provides a ranking of each feature based on its influence on the output label, allowing us to determine which dimension(s) have greater control over the specific edit. 
Having identified the relevant dimensions, the method performs the desired transformation by swapping the top-$K$ most essential dimensions with the corresponding dimensions from a reference image that exhibits the desired attribute.
This explains why our method is dubbed \emph{\model{}}.
After showing how the number $K$ of swapped dimensions controls the intensity of the transformation, we propose a method to automatically tune $K$ on a per-sample basis to apply the edit without excessively altering the input image, \eg{,} preserving the identity of the face. The project page can be found at \href{https://enisimsar.github.io/latentswap3d/}{\textit{https://enisimsar.github.io/latentswap3d/}}. Our contributions can be summarized as follows:
\begin{itemize}
    \item We explore 3D GAN latent spaces to determine their ability to encode semantic attributes in terms of disentanglement, completeness, and informativeness~(DCI).
    \item We propose \model{} enabling attribute editing tasks for \emph{any pre-trained} 2D or 3D generative model \emph{without the need to re-train or fine-tune the generators}. \model{} achieves state-of-the-art semantic attribute editing results in terms of \textit{semantic correctness} by preserving identity. 
    \item We first show results for attribute editing of generated images from random seeds of the 3D generators, then we broaden the capabilities of \model{} to edit the attributes of real images by applying a vanilla GAN inversion or off-the-shelf GAN inversion methods.
\end{itemize}

We test our method by applying the most popular and state-of-the-art generators: \pigan{}~\cite{pigan}, MVCGAN~\cite{zhang2022multi}, EG3D~\cite{chan2021efficient}, StyleSDF~\cite{or2021stylesdf}, GIRAFFE~\cite{Niemeyer2020GIRAFFE}, StyleNeRF~\cite{gu2021stylenerf}, VolumeGAN~\cite{xu2021volumegan}, and StyleGAN2~\cite{karras2019style}, trained on five public datasets: FFHQ~\cite{karras2019style}, CelebA~\cite{liu2015faceattributes}, AFHQ~\cite{choi2020starganv2}, CompCars~\cite{yang2015large}, and MetFaces~\cite{Karras2020ada}. The main paper focuses on the editing results for \pigan{}, MVCGAN and EG3D in the FFHQ, CelebA and AFHQ datasets. %

\section{Related Work}
\label{sec:related}
\noindent\textbf{Image editing in GANs.} StyleGAN generators~\cite{karras2019style,Karras2020stylegan2,Karras2021} are widely used to generate high-quality images by converting a random noise vector into a latent code that can encode semantically meaningful attributes~\cite{wu2021stylespace,simsar2023fantastic}. Image editing can then be implemented as manipulations of those latent codes, either supervised~\cite{shen2020interfacegan,abdal2021styleflow,goetschalckx2019ganalyze,shi2022semanticstylegan,hu2022style} or unsupervised~\cite{shen2021closed,yuksel2021latentclr,patashnik2021styleclip,voynov2020unsupervised}.
Supervised methods are based on annotated labels or pre-trained attribute classifiers to predict the presence of semantic attributes.
InterFaceGAN~\cite{shen2020interfacegan} learns hyperplanes in latent space, whereas StyleFlow~\cite{abdal2021styleflow} employs conditional normalizing flows.
Unsupervised approaches, instead, do not require pre-trained classifiers or labels. 
Semantic Factorization (SeFa)~\cite{shen2021closed} finds semantic directions by retrieving eigenvectors from a projection matrix by singular value decomposition, while LatentCLR~\cite{yuksel2021latentclr} uses a contrastive learning-based method to learn directions. 
Such editing methods are developed primarily for StyleGAN, which has special linearly editable latent spaces~\cite{wu2021stylespace}. However, many 3D GANs~\cite{pigan,zhang2022multi,Niemeyer2020GIRAFFE,or2021stylesdf,xu2021volumegan} use a non-linear style integration unit~\cite{perez2018film}, making direct 2D editing methods ineffective and causing unwanted effects such as identity change, degenerate facial attributes, and entangled edits.
In this work, we propose a generalizable semantic editing method that can be used with any 3D or 2D GAN. 

\noindent\textbf{3D GANs.} 
Recent advancements in combining NeRF with GAN have led to the development of 3D GANs~\cite{graf,pigan,zhang2022multi,gu2021stylenerf,pan2021shading,xu2021generative} that allow explicit control over the pose of the object being generated. There are two trends for the 3D GAN architectures: (i) \textbf{ one-staged:} use pure volumetric rendering in the generator and (ii) \textbf{two-staged:} use a combination of low-resolution volumetric rendering and 2D GANs to increase the output resolution.
GRAF~\cite{graf} and \pigan~\cite{pigan} are one-stage generators that provide 3D-aware image and geometry generation using an implicit neural rendering but cannot afford high resolution while training. 
Two-stage generators~\cite{gu2021stylenerf,Niemeyer2020GIRAFFE,zhang2022multi,or2021stylesdf,chan2021efficient}, include StyleNeRF~\cite{gu2021stylenerf} and MVCGAN~\cite{zhang2022multi}, which use NeRF-based 3D renderers, and StyleSDF~\cite{or2021stylesdf}, which employs Signed Distance Fields (SDF)-based 3D renderers as the first stage. Additionally, EG3D~\cite{chan2021efficient} introduces a hybrid explicit and implicit 3D representation through a tri-plane. Our work proposes an edit method that can be used with any of these models off-the-shelf without additional GAN training. 

\noindent\textbf{3D appearance \& shape edits.} Existing research on attribute editing methods for 3D shapes and appearances focuses mainly on learning an edit during the training phase. 
3D face generation methods~\cite {stylerig,mostgan,KowalskiECCV2020,shoshan2021gan_control,liu20223dfmgan,lee2022expgan,kwak2022injecting,tang2022explicitly,zhang2022training} often enable explicit control over attributes. 
However, some of those methods~\cite{liu20223dfmgan,mostgan,stylerig,tang2022explicitly,deng2020disentangled,tewari2020pie,sun2022controllable} are limited to editing only pose, expression, and illumination, while others~\cite{KowalskiECCV2020,shoshan2021gan_control,zhang2022training} use a set of predefined labels or losses during the training process, limiting controllability during generation.
One of them, CONFIG~\cite{KowalskiECCV2020}, is trained on real and synthetic data with predefined attributes from scratch to enable semantic editing.
Alternatively, 
\cite{sun2022fenerf,sun2022ide,chen2022sem2nerf,jiang2022nerffaceediting,deng2023pix2pix3d}
propose 3D generators that enable portrait image editing by utilizing semantic maps. However, they also require re-training of the generators from scratch.
\cite{3dgan_inversion} showed high-quality and disentangled edits, such as \textit{gender}, and \textit{age}, using StyleFlow~\cite{abdal2021styleflow} on pre-trained EG3D~\cite{chan2021efficient}. However, we will show how StyleFlow underperforms for other attributes and on other 3D GANs. Most of the methods above have a restricted focus, as they can only manipulate the attributes of portrait images and cannot be applied to other datasets~\cite{cats,choi2020starganv2,yang2015large}. Furthermore, these methods are not architecture agnostic and apply attribute editing as one of the tasks optimized during the training phase~\cite{KowalskiECCV2020,shoshan2021gan_control}.
Our method instead enables attribute editing on any generator without requiring GAN training and on any dataset, such as human faces, animals, or cars, as we show in experiments.

\noindent\textbf{Image inversion for generative models.} Editing on real images is possible by obtaining the latent code for an input image by \textit{GAN inversion}. There are different inversion approaches, from learning-based by using encoder networks~\cite{perarnau2016invertible,richardson2021encoding,tov2021designing} to optimization-based \cite{zhu2016generative,abdal2019image2stylegan} or hybrid \cite{bau2019seeing,zhu2020domain}. 
Several 3D-GAN inversion methods have recently been proposed, including optimization-based~\cite{pigan,xu2023inNout,yin2022nerfinvertor} and learning-based methods~\cite{unsupervised_novel_view,cai2022pix2nerf,ko20233d}. We also incorporate image inversion with \model{} for real image edits.

\section{\model}
\label{sec:method}
\subsection{Overview} \label{sec:overview}

\begin{figure*}[ht!]
    \centering
    
    \begin{subfigure}{.46\linewidth}
        \centering
        \includegraphics[width=\linewidth]{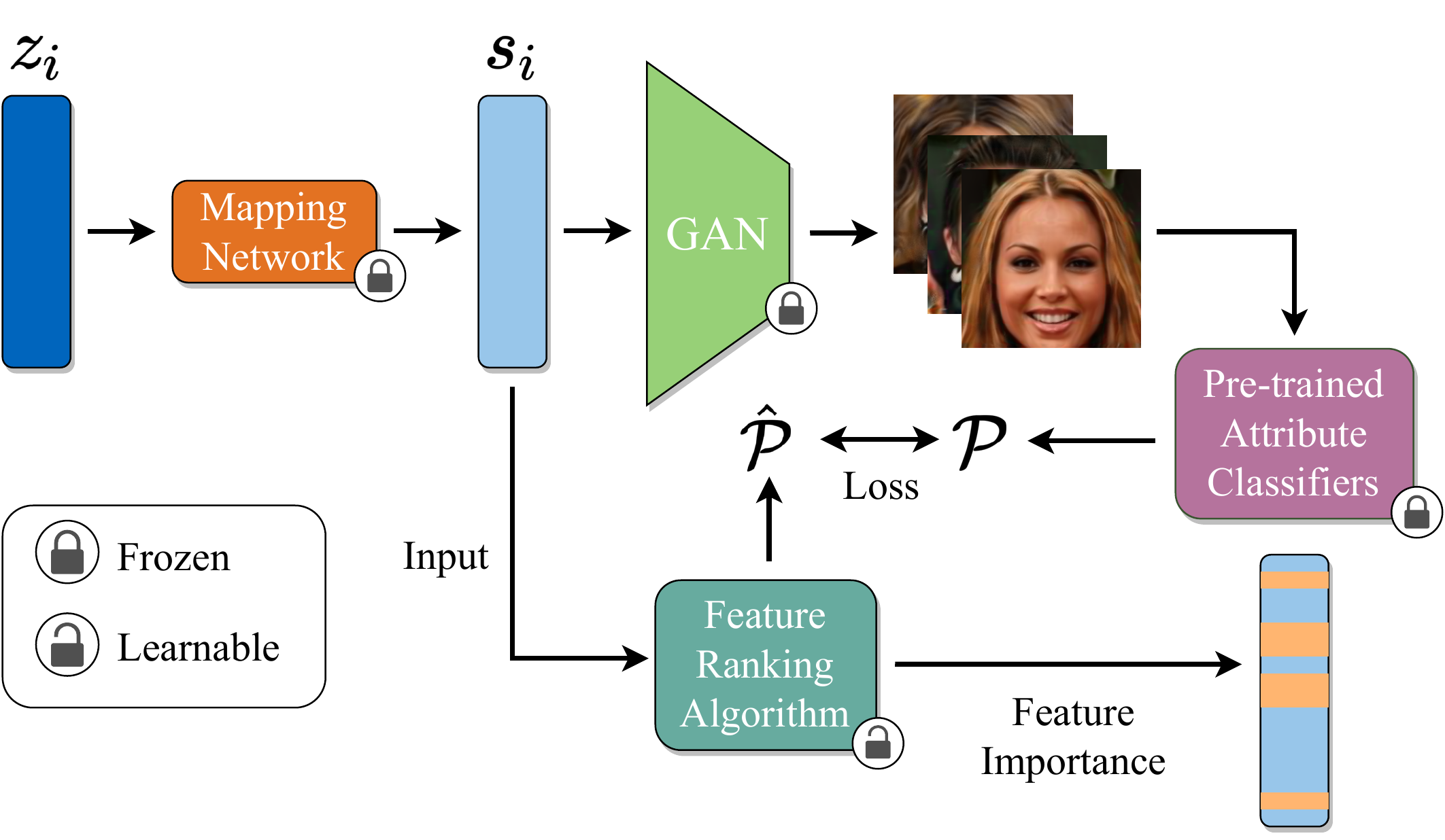}
        \caption{Identifying Relevant Latent Dimensions.}
        \label{fig:step1}
    \end{subfigure}
    \hfill\vline\hfill
    \begin{subfigure}{.50\linewidth}
        \centering
        \includegraphics[width=\linewidth]{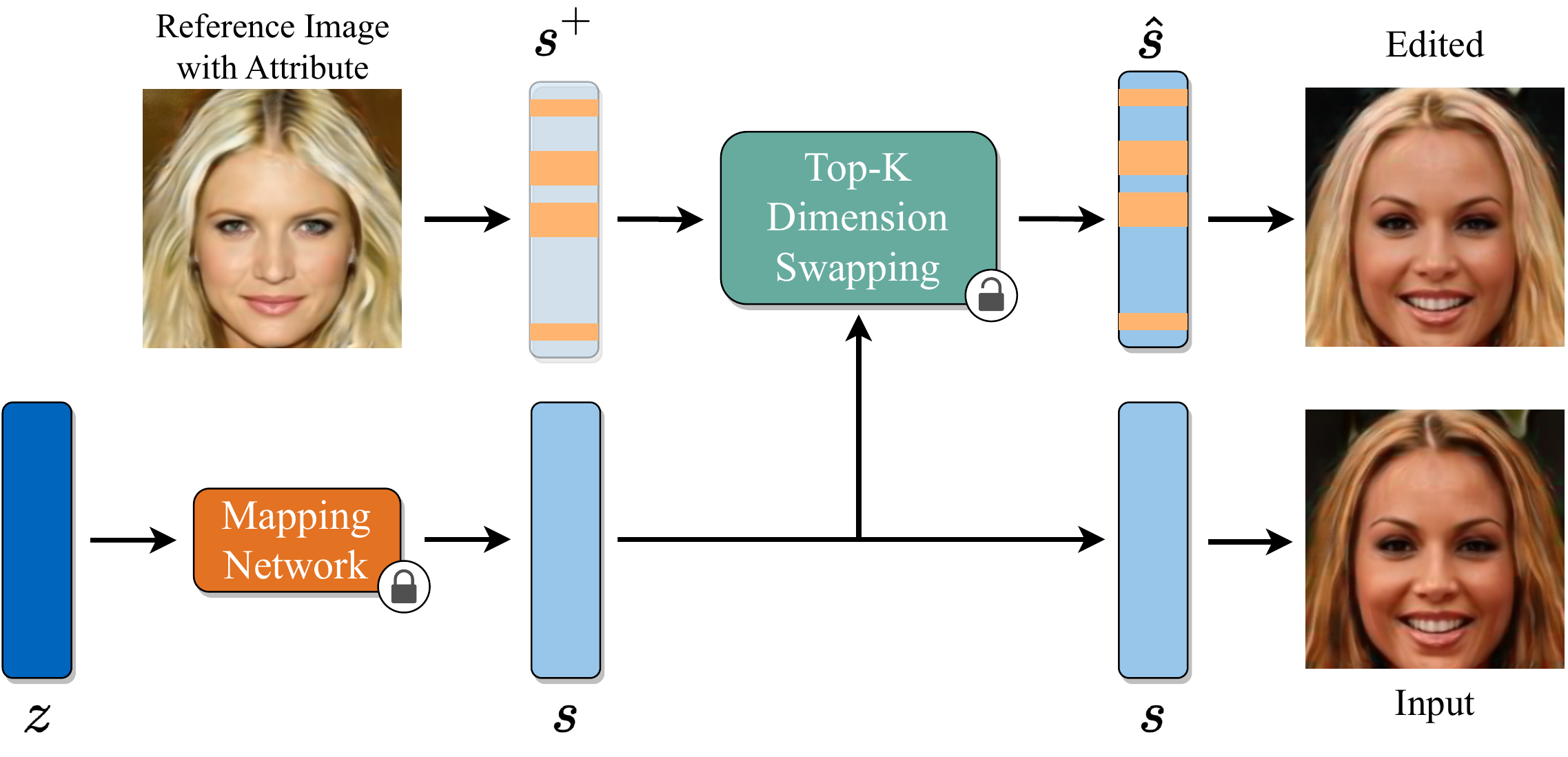}
        \caption{Attribute Editing on Latent Dimensions.}
        \label{fig:step2}
    \end{subfigure}
    
    \caption{(a) We propose to train a random forest regressor taking latent codes $s_i$ to predict the presence/absence of a desired attribute. We use the trained forest to rank the importance of dimensions of $s_i$ concerning the desired attribute. (b) Given the latent code $s$ of an image, first we find the closest latent code in the support set exhibiting the desired attribute (\eg{,} $s^+$ to increase \textit{ blondeness}), then we swap the top $K$ dimensions related to the attribute to generate an edited latent code $\hat{s}$ that can be decoded in an edited image.}
    \vspace{-0.2cm}
    \label{fig:framework}
\end{figure*}

We aim to build a generator-agnostic method for \emph{any pre-trained} 3D GAN \emph{without re-training or fine-tuning}.
\model{} consists of two main components. The first one identifies essential features in the latent space of a 3D GAN that controls the desired attribute through a random forest algorithm. Then, the target attribute is applied in an identity-preserving manner through a feature-swapping approach, see Fig.~\ref{fig:framework} for an overview of the two components.

\subsection{Background} \label{sec:background}

Neural radiance fields~(NeRFs) are represented as a set of multilayer perceptrons~(MLPs), taking as input a 3D coordinate ($\mathbf{x, y, z}$), and camera azimuth and elevation angles ($\phi, \theta$). The output is a spatially varying density and a viewpoint-dependent color. Finally, an image is rendered by sampling rays from the camera location towards the image plane and evaluating the related radiance values~\cite{mildenhall2020nerf}.

3D GANs are built on top of the exact volumetric rendering and aim to learn to generate NeRF-like volumes from a sampled latent noise vector by training only on unlabeled 2D images.
While for 2D GANs, such as StyleGAN\cite{Karras2020stylegan2}, the generation is controlled by Adaptive Instance normalization (AdaIN)~\cite{huang2017arbitrary}, for a popular family of 3D-GANs, \eg{,} \pigan{} or MVCGAN, it is controlled by feature-wise linear modulation (FiLM)~\cite{perez2018film} which learns functions $f$ and $h$ which output $\gamma_{i,c}$ and $\beta_{i,c}$ as a function of input $\bm{x_i}$, $\gamma_{i,c} = f_c(\bm{x}_i)$ and $\beta_{i,c} = h_c(\bm{x}_i)$ where $\gamma_{i,c}$ and $\beta_{i,c}$ modulate a neural network's activations $\bm{F}_{i,c}$ of $i^{th}$ input's $c^{th}$ feature map, via a feature-wise affine transformation:
\begin{equation} \label{eq:film}
  FiLM(\bm{F}_{i,c} | \gamma_{i,c}, \beta_{i,c}) = \gamma_{i,c}\bm{F}_{i,c} + \beta_{i,c}.
\end{equation}
$f$ and $h$ can be arbitrary functions such as neural networks. 
In this family of generators, $\bm{x}_i$ is the position in space to render, and $\gamma_{i,c}$ and $\beta_{i,c}$ are obtained starting from an input latent code $z$ and fed into a mapping network to guide image generation through SIREN~\cite{sitzmann2019srns} based FiLM layers: 
\begin{equation} \label{eq:film_siren}
    \phi_{i,c}(x_i) = \sin(FiLM(\bm{F}_{i,c} | \gamma_{i,c}, \beta_{i,c})).
\end{equation}
A notable exception to this sinusoidal modulation paradigm is represented by EG3D~\cite{chan2021efficient}, which inherits the network structure and modulation style of the StyleGAN family~\cite{Karras2020stylegan2} of generative models to generate three planes of features whose inner product defines the volume used for rendering.

For both families, the underlying idea of having a mapping network re-parametrizing the conditioning vector from a random multivariate normal distribution to a \textit{modulation/style space} is shared. However, for models like \pigan{} and MVCGAN, the use of sine activation functions makes the latent space periodic and, therefore, more challenging to control compared to models based on AdaIn layers (\eg{,} StyleGAN or EG3D). For instance, $\beta_{i,c}$ is the phase shift of a sine function and, according to Eq.~\ref{eq:film_siren}, will give the same output for every $\beta_{i,c} + 2k * \pi$ with $k \in \mathbb{Z}$. While $\gamma_{i,c}$ controls the frequency of the sine function and affects the periodicity of the output. In practice, linear increases or decreases of $(\beta_{i,c},\gamma_{i,c})$ might result in the opposite effect on the output of the sinusoidal activation. This causes some of the method proposals for the latent space of GANs based on AdaIN layers to fail, as shown in Sec.~\ref{sec:experiment}. Furthermore, all 3D GANs include several latent spaces, therefore, we need to identify the most suited one for attribute editing.

\subsection{Identifying Relevant Latent Dimensions} \label{sec:ranking}

The core idea of \model{} lies in using a feature ranking algorithm to determine the importance of features for a given attribute. In particular, for all experiments, we rely on a random forest~\cite{breiman2001random} due to its explainability. An overview of the process is summarized in Fig.~\ref{fig:step1}. To find relevant dimensions in the latent space of a 3D GAN, we start by generating a set of images from randomly sampled latent codes $z_i$ and corresponding mapped codes $s_i$. Then, we assign an attribute score $\mathcal{P}$ for each image in the generated set using pre-trained image attribute classifiers. The scores correspond to the presence/absence of a particular attribute in the generated images. Using these scores, we train a random forest classifier to predict the presence of an attribute from the latent codes of the generator. Since random forests are very effective models for ranking feature importance, we can explicitly identify the dimensions of the latent code that correspond to desired attributes. In practice, we use the occurrence with which a forest decision node selects the input dimensions to rank the relevance of each dimension regarding the presence of a specific attribute~\cite{criminisi2013decision}. 

\subsection{Attribute Editing on Latent Dimensions}\label{sec:editing}
The existing 2D GAN editing methods perform semantic editing on latent spaces by applying algebraic operations. However, this is not applicable to 3D GANs that utilize periodic activation functions during the style integration process, such as \pigan{} and MVCGAN, as discussed in Sec.~\ref{sec:background}, which is parameterized by a frequency and phase shift. 
Inspired by the style mixing method proposed in StyleGAN, we realize image editing by swapping dimensions between reference and target latent codes. While style mixing swaps entire blocks of latent codes to realize interpolation between two hand-picked latent codes, we automatically identify the target code and use the ranking identified in Sec.~\ref{sec:ranking} to precisely swap only a small subset of dimensions. Thanks to this targeted swap, we achieve edits that do not alter the identity of the original image. 

We demonstrate the attribute editing process in Fig.~\ref{fig:step2}.
After determining the ranking of latent dimensions for a given attribute with the random forest, \model{} replaces the top-$K$ features of the latent code of an image being manipulated ($s$) with those of an image ($s^+$) taken from the support set used to train the random forest and exhibiting the desired attribute. The output latent codes ($\hat{s}$) generate the edited image with the desired attribute. 
In particular, we pick a reference image whose attribute score is the lowest/highest for the desired attribute to remove/add the corresponding transformation to the manipulated image.

The parameter $K$ should be carefully chosen for each transformation to preserve the identity of the generated image after attribute editing. We use the identity loss $\mathcal{L}_{ID}$ presented in Encoder4Editing~\cite{tov2021designing} to automatically tune the parameter $K$ on a per-sample basis. This loss calculates the cosine similarity between the feature embedding of the original image and that of the edited image. For example, in the face domain, we compute $\mathcal{L}_{ID}$ based on a pre-trained ArcFace~\cite{deng2018arcface} face recognition network, while for other domains, a ResNet-50~\cite{resnet50} network trained for MOCOv2~\cite{chen2020improved} is used. In particular, we select the maximum $K$ that satisfies the constraint $\mathcal{L}_{ID} < \tau$. We provide ablation studies of the parameter $K$ in Sec.~\ref{sec:qualitative}. %

To maximize identity preservation, we also choose a suitable reference image by: first selecting the top $N$ images with the highest attribute score from the support set; then choosing the most similar to the one currently being edited according to the cosine similarity between the respective latent codes. This process ensures that features will be swapped among similar samples sharing most attributes except the one we would like to modify.

\subsection{3D Attribute Edits on Real Images} 
\label{sec:editonreal}
Applying \model{} to a real image requires first GAN inversion~\cite{xia2021gan} to embed it in the latent space of the pre-trained GAN generator. Furthermore, the inversion of 3D GANs also requires finding the camera pose from which the real image has been acquired~\cite{unsupervised_novel_view}.

\begin{figure}[ht!]
    \centering
    \includegraphics[width=\linewidth]{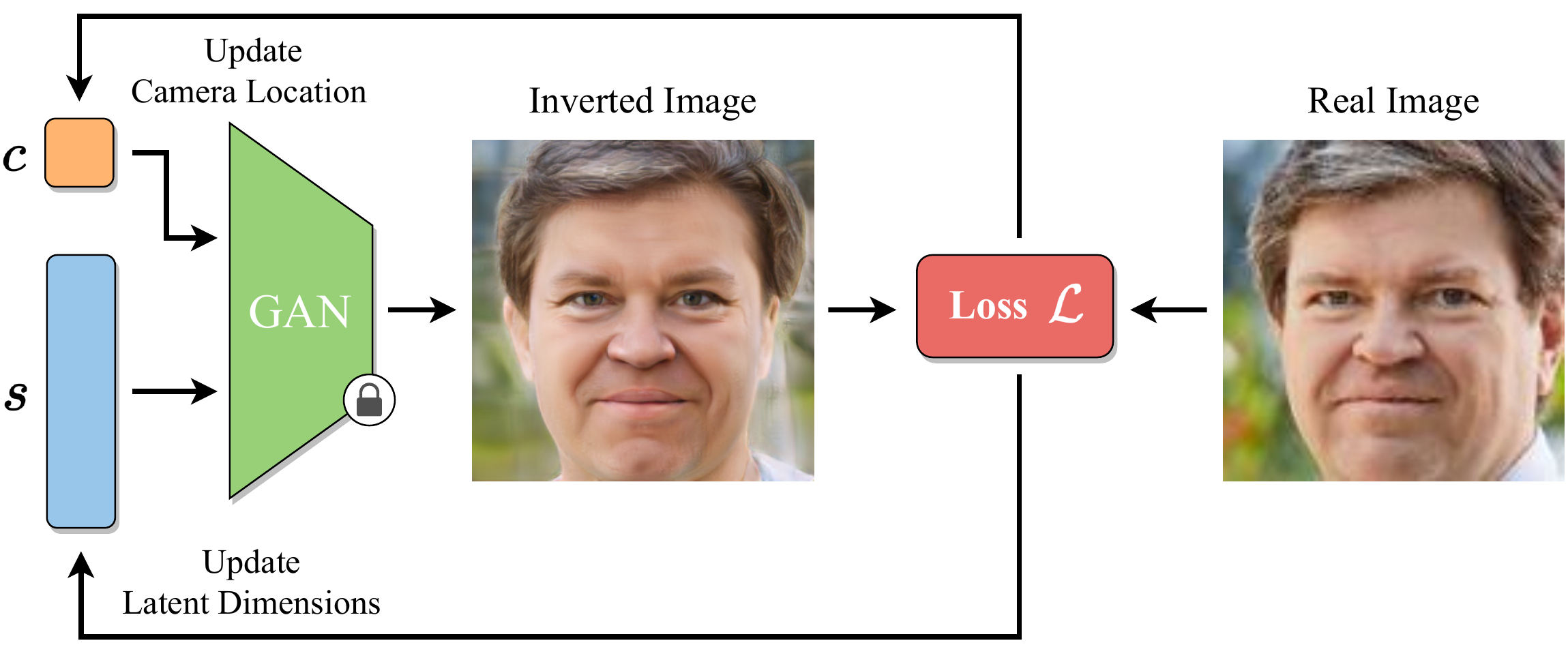}
    \caption{Inversion pipeline for a real image. $s, c$ corresponds to the latent code and camera location, respectively.}
    \label{fig:framework_inversion}
\end{figure}

For 3D GAN inversion, we follow an iterative optimization approach summarized in Fig.~\ref{fig:framework_inversion}, where the latent vector and pose are optimized alternatively. First, the latent vector is initialized to the mean vector in the latent space, while the camera pose is initialized to a neutral frontal position. Next, we start the inversion process by optimizing the camera location, $c$, while freezing the latent code, then we swap roles and tune the latent vector $s$, keeping the camera fixed. This process is repeated for a number of optimization steps. Then the camera is fixed, while the latent code is further optimized for a fixed number of steps.

We use a linear combination of reconstruction losses computed between the generated image and the reference one to guide the optimization: $\mathcal{L}_{2}$, $\mathcal{L}_{LPIPS}$~\cite{zhang2018perceptual}, and identity Loss $\mathcal{L}_{ID}$~\cite{tov2021designing}:
\begin{equation} \label{eq:inversion}
    \mathcal{L} = \lambda_{1}\mathcal{L}_{2} + \lambda_{2}\mathcal{L}_{LPIPS} + \lambda_{3}\mathcal{L}_{ID}
\end{equation}
where the values of $\lambda$s are specified in Sec.~\ref{sec:experiment}.

\begin{figure*}[ht!]
    \centering
    \begin{minipage}{.33\linewidth}
        \centering
        \textbf{\pigan{}}
    \end{minipage}
    \begin{minipage}{.33\linewidth}
        \centering
        \textbf{MVCGAN}
    \end{minipage}
    \begin{minipage}{.33\linewidth}
        \centering
        \textbf{EG3D}
    \end{minipage}
    
    \vspace{0.1cm}

    \begin{minipage}{.33\linewidth}
        \centering
        \includegraphics[width=\linewidth]{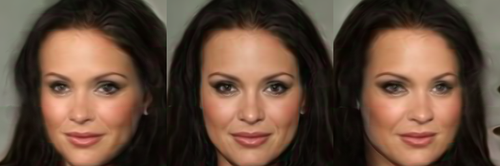}
    \end{minipage}
    \begin{minipage}{.33\linewidth}
        \centering
        \includegraphics[width=\linewidth]{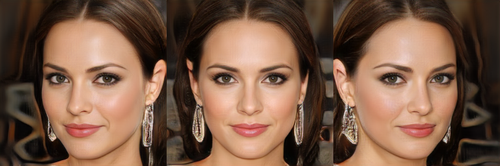}
    \end{minipage}
    \begin{minipage}{.33\linewidth}
        \centering
        \includegraphics[width=\linewidth]{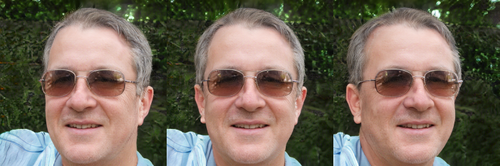}
    \end{minipage}

    \begin{minipage}{.2439\linewidth}
        \centering
        \includegraphics[width=\linewidth]{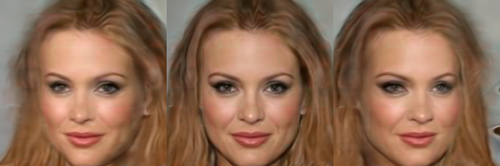}
    \end{minipage}
    \begin{minipage}{.0813\linewidth}
        \centering
        \includegraphics[width=\linewidth]{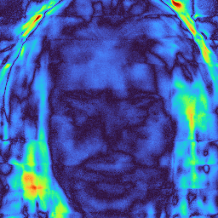}
    \end{minipage}
    \begin{minipage}{.2439\linewidth}
        \centering
        \includegraphics[width=\linewidth]{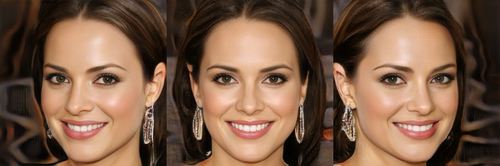}
    \end{minipage}
    \begin{minipage}{.0813\linewidth}
        \centering
        \includegraphics[width=\linewidth]{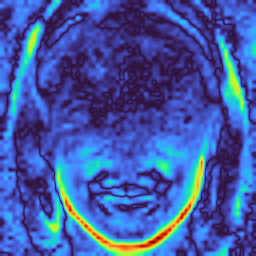}
    \end{minipage}
    \begin{minipage}{.2439\linewidth}
        \centering
        \includegraphics[width=\linewidth]{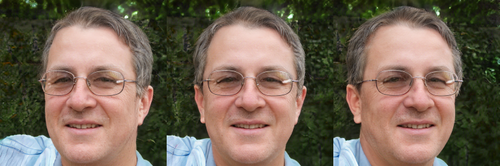}
    \end{minipage}
    \begin{minipage}{.0813\linewidth}
        \centering
        \includegraphics[width=\linewidth]{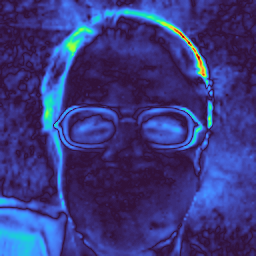}
    \end{minipage}
    
    \begin{minipage}{.2439\linewidth}
        \centering
        \textbf{\scriptsize{Blonde (+) \phantom{g}}}
    \end{minipage}
    \begin{minipage}{.0813\linewidth}
        \centering
        \phantom{depth}
    \end{minipage}
    \begin{minipage}{.2439\linewidth}
        \centering
        \textbf{\scriptsize{Smiling (+)}}
    \end{minipage}
    \begin{minipage}{.0813\linewidth}
        \centering
        \phantom{depth}
    \end{minipage}
    \begin{minipage}{.2439\linewidth}
        \centering
        \textbf{\scriptsize{Eyeglasses (-)}}
    \end{minipage}
    \begin{minipage}{.0813\linewidth}
        \centering
        \phantom{depth}
    \end{minipage}
    
    \begin{minipage}{.2439\linewidth}
        \centering
        \includegraphics[width=\linewidth]{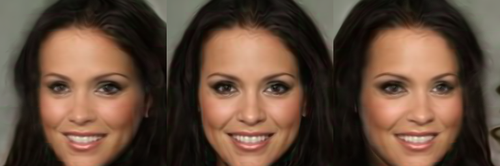}
    \end{minipage}
    \begin{minipage}{.0813\linewidth}
        \centering
        \includegraphics[width=\linewidth]{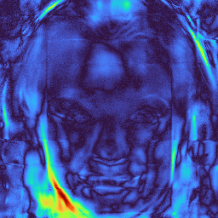}
    \end{minipage}
    \begin{minipage}{.2439\linewidth}
        \centering
        \includegraphics[width=\linewidth]{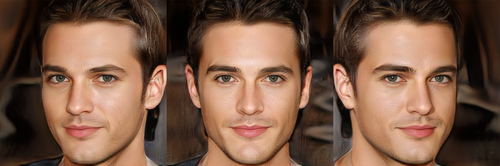}
    \end{minipage}
    \begin{minipage}{.0813\linewidth}
        \centering
        \includegraphics[width=\linewidth]{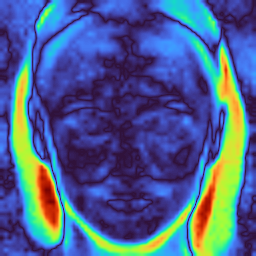}
    \end{minipage}
    \begin{minipage}{.2439\linewidth}
        \centering
        \includegraphics[width=\linewidth]{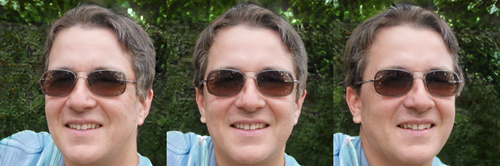}
    \end{minipage}
    \begin{minipage}{.0813\linewidth}
        \centering
        \includegraphics[width=\linewidth]{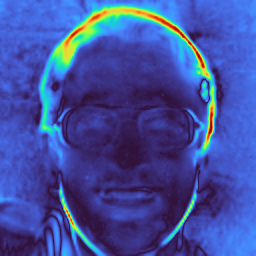}
    \end{minipage}
    
    \begin{minipage}{.2439\linewidth}
        \centering
        \textbf{\scriptsize{Smiling (+) \phantom{g}}}
    \end{minipage}
    \begin{minipage}{.0813\linewidth}
        \centering
        \phantom{depth}
    \end{minipage}
    \begin{minipage}{.2439\linewidth}
        \centering
        \textbf{\scriptsize{Male (+)}}
    \end{minipage}
    \begin{minipage}{.0813\linewidth}
        \centering
        \phantom{depth}
    \end{minipage}
    \begin{minipage}{.2439\linewidth}
        \centering
        \textbf{\scriptsize{Age (-)}}
    \end{minipage}
    \begin{minipage}{.0813\linewidth}
        \centering
        \phantom{depth}
    \end{minipage}

    \begin{minipage}{.2439\linewidth}
        \centering
        \includegraphics[width=\linewidth]{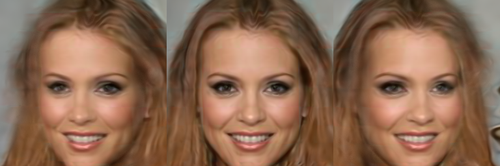}
    \end{minipage}
    \begin{minipage}{.0813\linewidth}
        \centering
        \includegraphics[width=\linewidth]{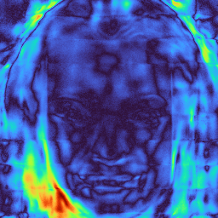}
    \end{minipage}
    \begin{minipage}{.2439\linewidth}
        \centering
        \includegraphics[width=\linewidth]{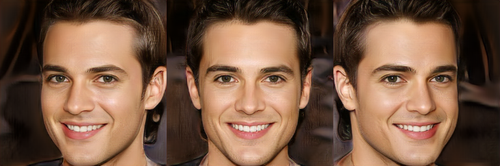}
    \end{minipage}
    \begin{minipage}{.0813\linewidth}
        \centering
        \includegraphics[width=\linewidth]{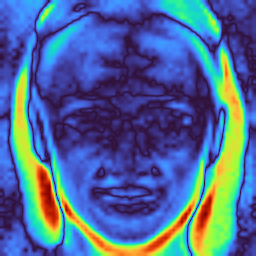}
    \end{minipage}
    \begin{minipage}{.2439\linewidth}
        \centering
        \includegraphics[width=\linewidth]{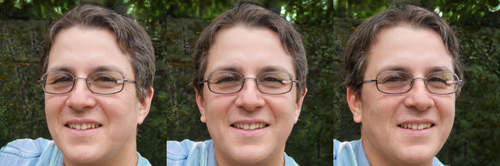}
    \end{minipage}
    \begin{minipage}{.0813\linewidth}
        \centering
        \includegraphics[width=\linewidth]{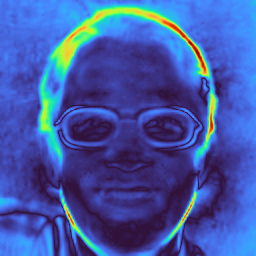}
    \end{minipage}
    
    \begin{minipage}{.2439\linewidth}
        \centering
        \textbf{\scriptsize{Blonde (+) \& Smiling (+) \phantom{g}}}
    \end{minipage}
    \begin{minipage}{.0813\linewidth}
        \centering
        \phantom{depth}
    \end{minipage}
    \begin{minipage}{.2439\linewidth}
        \centering
        \textbf{\scriptsize{Smiling (+) \& Male (+)}}
    \end{minipage}
    \begin{minipage}{.0813\linewidth}
        \centering
        \phantom{depth}
    \end{minipage}
    \begin{minipage}{.2439\linewidth}
        \centering
        \textbf{\scriptsize{Eyeglasses (-) \& Age (-)}}
    \end{minipage}
    \begin{minipage}{.0813\linewidth}
        \centering
        \phantom{depth}
    \end{minipage}

\caption{Qualitative results for our method, \model{}, on CelebA \cite{liu2015faceattributes} with \pigan{} \cite{pigan}, CelebA~\cite{liu2015faceattributes} with MVCGAN~\cite{zhang2022multi} and FFHQ~\cite{karras2019style} with EG3D~\cite{chan2021efficient}. A random seed image from different viewpoints is shown on the first row, followed by the edits for specific attributes and their combination (cumulative edits) in the last row, where (+/-) indicates an increase/decrease in the edited attribute. The rightmost column shows a heatmap of the changes in the underlying 3D geometry between the edited and original image.}

\label{fig:results}
\end{figure*}

\section{Experiments}
\label{sec:experiment}
\model{} is tested with three state-of-the-art 3D GANs: \pigan~\cite{pigan} as a representative of one-stage 3D generators with periodical latent spaces, MVCGAN~\cite{zhang2022multi} as a representative of two-stage 3D generators with high fidelity output, and EG3D~\cite{chan2021efficient} used to show how our proposal also generalizes to generators that inherit the latent spaces of StyleGAN. The images used in the experiments are from four different datasets: Flickr-Faces-HQ~(FFHQ)~\cite{karras2019style}, Large-scale CelebFaces Attributes~(CelebA)~\cite{liu2015faceattributes} Cats~\cite{cats} and Animal Faces-HQ~(AFHQ)~\cite{choi2020starganv2}.

\textbf{\pigan{} \cite{pigan}} is a NeRF-based one-staged 3D GAN. A random noise vector $z \in \mathcal{Z}$ is first transformed into a 4608-dimensional vector $s \in \mathcal{S}$, corresponding to the frequency and phase shifts of FiLM layers. For \pigan{}, we use this $\mathcal{S}$ space to apply our edits.

\textbf{MVCGAN~\cite{zhang2022multi}} proposes a two-stage 3D GAN. In the first stage, a neural volume renderer generates a low-resolution image and the geometry of a shape. In the second stage, a 2D styles-based generator enables high-resolution image generation. Its mapping network is converting from random noise $z \in \mathcal{Z}$ into intermediate latent codes $s \in \mathcal{S}$, conditioning the neural rendered. $\mathcal{S}$ Space has 4864 dimensions and is the space we select to apply our edits.

\textbf{EG3D~\cite{chan2021efficient}} is also a two-stage 3D GAN. Unlike MVCGAN, EG3D firstly feeds latent codes to a style-based 2D generator that predicts three orthogonal feature planes corresponding to the $x, y, z$ axes of a 3D volume. Then, it uses a neural volumetric renderer to decode interpolated features from the three planes into a low-resolution image that later gets fed to a 2D super-resolution network. The mapping network of EG3D converts random noise $z \in \mathcal{Z}$ into intermediate latent codes $s \in \mathcal{S}$, which has 7168 dimensions and is the space we select to apply our edits.

\textbf{Implementation.} \model{} is investigated on $10K$ synthesized images for each dataset to train random forests. Ablation on the size set can be found in \emph{Supplementary Material}. For the face attributes, such as \textit{gender}, \textit{age}, and \textit{hair color}, we use the pre-trained attribute models of the StyleGAN~\cite{Karras2020stylegan2} linear separability metrics. For both Cats and AFHQ datasets, we train a model for each attribute using annotated data from PetFinder.my Adoption Prediction Dataset~\cite{petfinder}.
For the selection of the number of features $K$, we set $\tau$ to $0.25$ for the face domain and $0.1$ for other domains. 
During the attribute editing step, we use a support set of 32 images among whom to pick the reference image. The ablation study on choosing $\tau$ and support set size can also be found in \emph{Supplementary Material}.
The weights in Eq.~\ref{eq:inversion} are tuned to $\lambda_1=1.0$, $\lambda_2=0.6$ and $\lambda_3=0.3$. To rank feature importance, we use the mean decrease in impurity and the Scikit-learn~\cite{scikit-learn} implementation. 

\subsection{Exploration of 3D GAN Latent Space} \label{sec:disentaglement}

There has been limited investigation into the exploration of the latent spaces of 3D GANs. As a result, the initial phase of our research involves assessing the disentanglement, completeness, and informativeness of a latent space by utilizing the DCI metrics proposed in~\cite{eastwood2018framework} and adapted in StyleSpace~\cite{wu2021stylespace}. We conduct experiments to explore each generator's most suitable latent space. 

\begin{table}[!ht]
  \centering
  \begin{tabular}{|l|c|ccc|}
    \hline
    \textbf{Generator} & \textbf{Space} & \textbf{Disent.$\uparrow$} & \textbf{Compl.$\uparrow$}  &\textbf{Inform.$\uparrow$}\\
    \hline
    \multirow{2}{*}{\pigan{}} & $\mathcal{Z}$ & 0.44 & 0.31 & 0.73\\ 
    & $\mathcal{S}$ & \textbf{0.80} & \textbf{0.91} & \textbf{0.98}\\
    \hline
    \multirow{2}{*}{MVCGAN} & $\mathcal{Z}$ & 0.43 & 0.30 & 0.75\\ 
    & $\mathcal{S}$ & \textbf{0.85} & \textbf{0.91} & \textbf{0.97}\\ 
    \hline
    \multirow{2}{*}{EG3D} & $\mathcal{Z}$ & 0.57 & 0.33 & 0.65 \\ 
    & $\mathcal{W}$ & \textbf{0.86} & \textbf{0.51} &\textbf{0.91}\\
    \hline
  \end{tabular}

  \caption{\label{tab:disent} DCI metrics for the different latent spaces of \pigan{}-CelebA, MVCGAN-CelebAHQ and EG3D-FFHQ. $\mathcal{Z}$ contains vectors sampled from a multivariate normal distribution. $\mathcal{S}$ and $\mathcal{W}$ represent the intermediate latent space.
  }
\end{table}

The training data of the DCI regressors are generated using 40 binary classifiers trained with the CelebA attributes~\cite{liu2015faceattributes} such as \textit{blond hair}, \textit{gender}, and \textit{eyeglasses}. $10K$ random noise vectors, $z \in \mathcal{Z}$, are sampled from a multivariate normal distribution and fed into the corresponding generator to get latent codes and the generated images used to train the DCI regressors. Table~\ref{tab:disent} shows how for \pigan{}, the $\mathcal{S}$ space has significantly better values in terms of disentanglement, completeness, and informativeness. MVCGAN shares similar latent spaces to \pigan{}, and the $\mathcal{S}$ space is better than the $\mathcal{Z}$ space. Finally, the DCI metrics for latent spaces of EG3D show that $\mathcal{W}$ space is better than the initial latent space $\mathcal{Z}$. This indicates that intermediate spaces of these models better disentangle the attributes of the generated images.

\subsection{Qualitative Evaluation} \label{sec:qualitative}

\noindent \textbf{Edits on generated images.} Figure~\ref{fig:results} illustrates qualitative edits on the CelebA dataset for \pigan{} and MVCGAN, and FFHQ for EG3D, where we apply manipulations on attributes such as \textit{blondness}, \textit{smiling}, \textit{changing gender}, \textit{eyeglasses type}, and \textit{age}. For these experiments, we sample a seed image from a frontal viewpoint, extract the latent code, and apply semantic edit in the latent space. Finally, we render the edited face from multiple views. We observe that our method indeed enables attribute edits in a disentangled fashion while maintaining 3D consistency from multiple views.

\begin{figure}[th!]
\centering
\vspace{-0.15cm}

\scriptsize
\begin{minipage}{.05\linewidth}
\rotatebox[origin=lc]{90}{\centering \hspace{-0.35cm}
\phantom{\scriptsize{\textbf{k}}}}
\end{minipage}
\begin{minipage}{.16\linewidth}
    \centering
    \textbf{Input}
\end{minipage}
\begin{minipage}{.48\linewidth}
    \centering
    \textbf{Breed}
\end{minipage}

\begin{minipage}{.05\linewidth}
    \raggedleft
\rotatebox[origin=lc]{90}{\scriptsize{\textbf{MVCGAN}}}
\end{minipage}
\begin{minipage}{.16\linewidth}
    \centering
    \includegraphics[width=\linewidth]{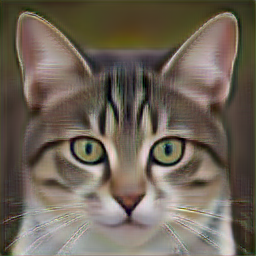}
\end{minipage}
\begin{minipage}{.48\linewidth}
    \centering
    \includegraphics[width=\linewidth]{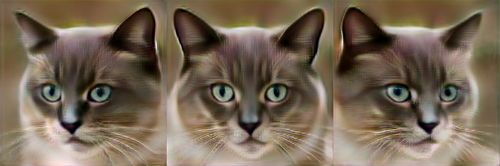}
\end{minipage}

\vspace{0.03cm}

\rotatebox[origin=lc]{90}{\centering \hspace{-0.35cm}
\phantom{\scriptsize{\textbf{k}}}}
\begin{minipage}{.7\linewidth}
    \centering
    \hrule width\linewidth\relax
\end{minipage}

\vspace{-0.03cm}

\begin{minipage}{.05\linewidth}
    \raggedleft
\rotatebox[origin=lc]{90}{\scriptsize{\textbf{EG3D}}}
\end{minipage}
\begin{minipage}{.16\linewidth}
    \centering
    \includegraphics[width=\linewidth]{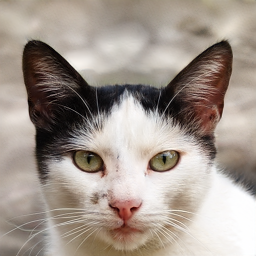}
\end{minipage}
\begin{minipage}{.48\linewidth}
    \centering
    \includegraphics[width=\linewidth]{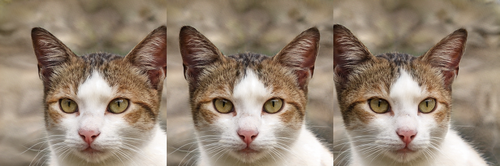}
\end{minipage}

\begin{minipage}{.05\linewidth}
\rotatebox[origin=lc]{90}{\centering \hspace{-0.35cm}
\phantom{\scriptsize{\textbf{k}}}}
\end{minipage}
\begin{minipage}{.16\linewidth}
    \centering
    \textbf{Input}
\end{minipage}
\begin{minipage}{.48\linewidth}
    \centering
    \textbf{Color}
\end{minipage}

\vspace{-0.2cm}

\caption{Results for 
AFHQ dataset~\cite{choi2020starganv2} with MVCGAN and EG3D generators. 
}
\label{fig:results_cats}
\end{figure}

We furthermore visualize 3D difference maps to evaluate the 3D consistency by extracting the depth maps from the underlying 3D geometry between the edited face and the original one and calculating the absolute depth differences. The rightmost column of Fig.~\ref{fig:results} shows difference maps in the form of heat maps, and the red color indicates distinct changes. Especially in the case of MVCGAN, visual edits correspond nicely to actual edits in the underlying 3D geometry (\eg{,} the smiling edit modifies the chin and lips). We further observe that the semantic edit quality is naturally bounded by the 3D generator's quality (\eg{,} difference maps from \pigan{} blonde and smiling are noisier).

In addition to disentangled attribute editing experiments on human faces, Fig.~\ref{fig:results_cats} shows the results of 
MVCGAN, and EG3D on the 
AFHQ dataset to prove the applicability of our method. Our method can successfully modify the breed and fur color.

\begin{figure}[ht!]
\centering
\scriptsize

\begin{minipage}{.2\linewidth}
    \centering
    \textbf{Input}
\end{minipage}
\begin{minipage}{.6\linewidth}
    \centering
    \textbf{Inverted}
\end{minipage}

\begin{minipage}{.2\linewidth}
    \centering
    \includegraphics[width=\linewidth]{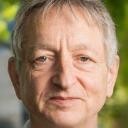}
\end{minipage}
\begin{minipage}{.6\linewidth}
    \centering
    \includegraphics[width=\linewidth]{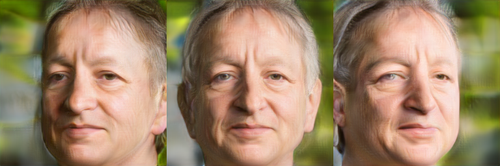}
\end{minipage}

\begin{minipage}{.2\linewidth}
    \centering
    \scriptsize{\textbf{Smiling}}
\end{minipage}
\begin{minipage}{.6\linewidth}
    \centering
    \includegraphics[width=\linewidth]{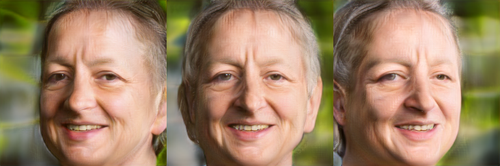}
\end{minipage}

\begin{minipage}{.2\linewidth}
    \centering
    \scriptsize{\textbf{Eyeglasses}}
\end{minipage}
\begin{minipage}{.6\linewidth}
    \centering
    \includegraphics[width=\linewidth]{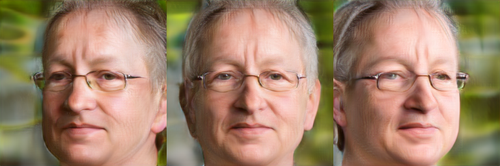}
\end{minipage}

\caption{Inversion, editing, and novel view synthesis for real images using our method and MVCGAN as a generator.}
\label{fig:inversion}

\end{figure}

\noindent\textbf{Edits on real images.} As explained in Sec.~\ref{sec:editonreal}, our method operates on real images captured from any viewpoint, then successfully performs editing tasks on them. Figure~\ref{fig:inversion} shows semantic edits, \ie{,} smiling and wearing eyeglasses, on the sample inverted in the latent space of MVCGAN.

\noindent\textbf{Comparison of CONFIG and \model{}.} 
We compare our proposed method, \model{}, to CONFIG~\cite{KowalskiECCV2020}, which is a neural face image generator developed to enable semantic edits. CONFIG has been explicitly trained to manipulate certain attributes and it requires a high amount of synthetic data, while \model{} finds the latent codes that enable the semantically meaningful edits on images without re-training the generator part.

\begin{figure}[ht!]
\centering
\scriptsize

\begin{minipage}{.13\linewidth}
    \centering
    \textbf{Input}
\end{minipage}
\begin{minipage}{.05\linewidth}
    \centering
    \phantom{\textbf{y}}
\end{minipage}
\begin{minipage}{.39\linewidth}
    \centering
    \textbf{CONFIG}
\end{minipage}
\begin{minipage}{.39\linewidth}
    \centering
    \textbf{\model{}}
\end{minipage}

\begin{minipage}{.13\linewidth}
    \centering
    \includegraphics[width=\linewidth]{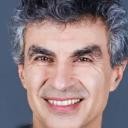}
\end{minipage}
\begin{minipage}{.03\linewidth}
    \raggedleft
    \rotatebox[origin=lc]{90}{\centering \hspace{-0.25cm} \scriptsize{\textbf{\phantom{y}Inverted}}}
\end{minipage}
\begin{minipage}{.39\linewidth}
    \centering
    \includegraphics[width=\linewidth]{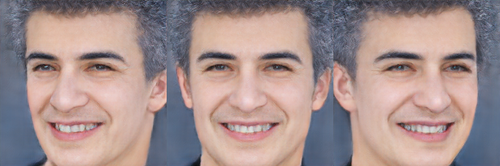}
\end{minipage}
\begin{minipage}{.39\linewidth}
    \centering
    \includegraphics[width=\linewidth]{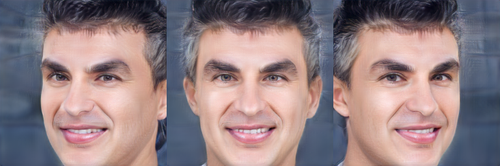}
\end{minipage}

\begin{minipage}{.13\linewidth}
    \centering
    \phantom{\includegraphics[width=\linewidth]{figures/comparison/confignet/bengio/gt.jpg}}
\end{minipage}
\begin{minipage}{.03\linewidth}
    \raggedleft
    \rotatebox[origin=lc]{90}{\centering \hspace{-0.15cm} \scriptsize{\textbf{Smiling}}}
\end{minipage}
\begin{minipage}{.39\linewidth}
    \centering
    \includegraphics[width=\linewidth]{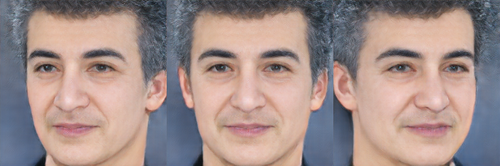}
\end{minipage}
\begin{minipage}{.39\linewidth}
    \centering
    \includegraphics[width=\linewidth]{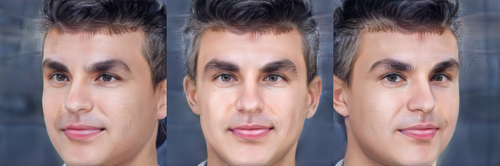}
\end{minipage}

\begin{minipage}{.13\linewidth}
    \centering
    \phantom{\includegraphics[width=\linewidth]{figures/comparison/confignet/bengio/gt.jpg}}
\end{minipage}
\begin{minipage}{.03\linewidth}
    \raggedleft
    \rotatebox[origin=lc]{90}{\centering \hspace{-0.15cm} \scriptsize{\textbf{Beard}}}
\end{minipage}
\begin{minipage}{.39\linewidth}
    \centering
    \includegraphics[width=\linewidth]{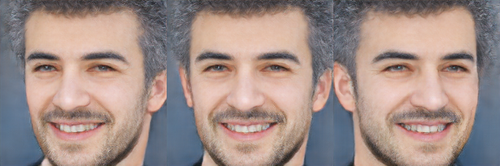}
\end{minipage}
\begin{minipage}{.39\linewidth}
    \centering
    \includegraphics[width=\linewidth]{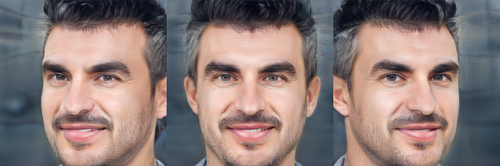}
\end{minipage}

\caption{Comparison of CONFIG~\cite{KowalskiECCV2020} and our method on smiling and beard attributes for a real face image.}
\label{fig:confignet_comparison}
\end{figure}

Figure~\ref{fig:confignet_comparison} shows real images attribute editing of the two methods for smiling (+) and beard. First, we observe that the inversion quality of real images in \model{} is better than the CONFIG method. Moreover, as the realism of semantic edits is tightly coupled to inversion quality, our \model{} generates more realistic edited images that preserve the identity of the subject. This experiment clearly shows the advantage of having a generator-agnostic method like \model{} that can easily harvest the latest advances on 3D consistent image generation over methods like CONFIG, which are bounded to a specific architecture and training regime.

\begin{figure}[ht!]
\centering

\scriptsize
\rotatebox[origin=lc]{90}{\centering \hspace{-0.45cm} \scriptsize{\textbf{\phantom{a}}}}
\begin{minipage}{.15\linewidth}
    \centering
    \textbf{Input}
\end{minipage}
\begin{minipage}{.15\linewidth}
    \centering
    \textbf{SeFa}
\end{minipage}
\begin{minipage}{.15\linewidth}
    \centering
    \textbf{LCLR.}
\end{minipage}
\begin{minipage}{.15\linewidth}
    \centering
    \textbf{IGAN.}
\end{minipage}
\begin{minipage}{.15\linewidth}
    \centering
    \textbf{StyleFlow}
\end{minipage}
\begin{minipage}{.15\linewidth}
    \centering
    \textbf{Ours}
\end{minipage}
\rotatebox[origin=lc]{90}{\centering \hspace{-0.45cm} \scriptsize{\textbf{\phantom{a}}}}

\rotatebox[origin=lc]{90}{\centering \hspace{-0.45cm} \scriptsize{\textbf{\pigan{}}}}
\begin{minipage}{.15\linewidth}
    \centering
    \includegraphics[width=\linewidth]{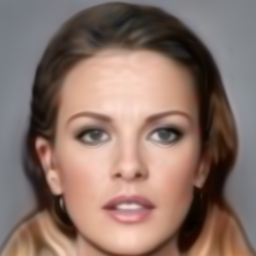}
\end{minipage}
\begin{minipage}{.15\linewidth}
    \centering
    \includegraphics[width=\linewidth]{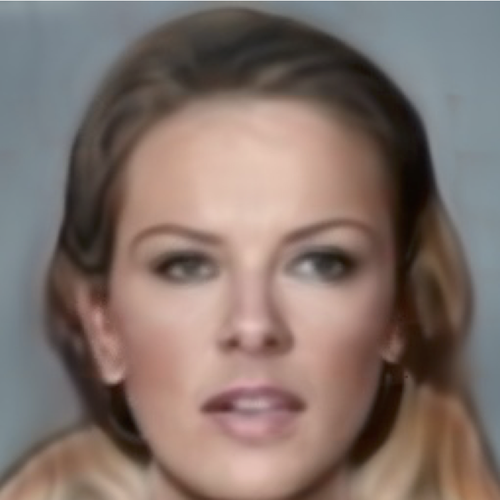}
\end{minipage}
\begin{minipage}{.15\linewidth}
    \centering
    \includegraphics[width=\linewidth]{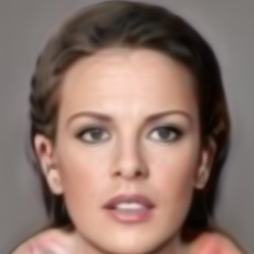}
\end{minipage}
\begin{minipage}{.15\linewidth}
    \centering
    \includegraphics[width=\linewidth]{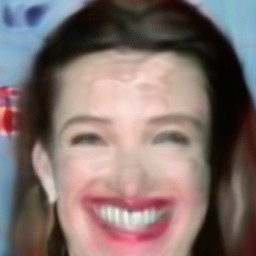}
\end{minipage}
\begin{minipage}{.15\linewidth}
    \centering
    \includegraphics[width=\linewidth]{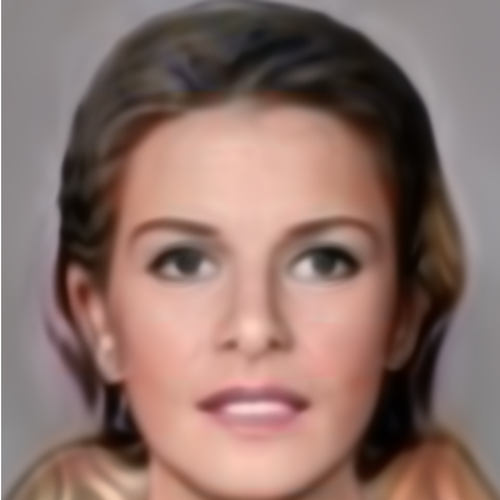}
\end{minipage}
\begin{minipage}{.15\linewidth}
    \centering
    \includegraphics[width=\linewidth]{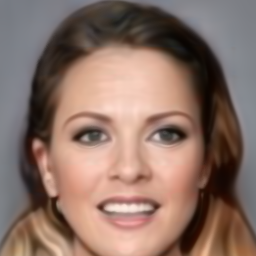}
\end{minipage}
\rotatebox[origin=lc]{90}{\centering \hspace{-0.63cm} \scriptsize{\textbf{Smiling (+)}}}

\rotatebox[origin=lc]{90}{\centering \hspace{-0.65cm} \scriptsize{\textbf{MVCGAN}}}
\begin{minipage}{.15\linewidth}
    \centering
    \includegraphics[width=\linewidth]{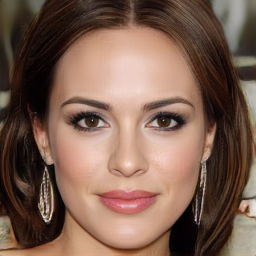}
\end{minipage}
\begin{minipage}{.15\linewidth}
    \centering
    \includegraphics[width=\linewidth]{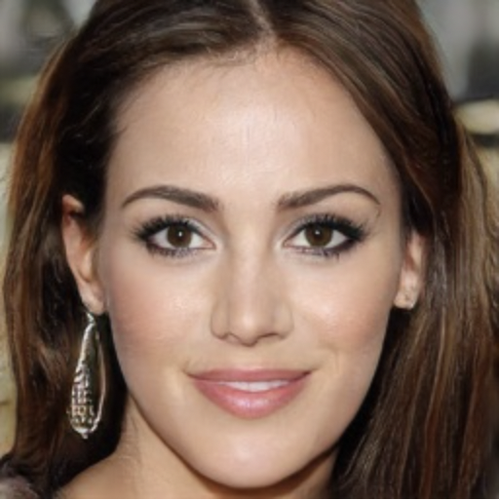}
\end{minipage}
\begin{minipage}{.15\linewidth}
    \centering
    \includegraphics[width=\linewidth]{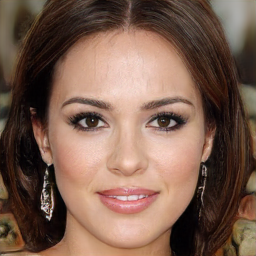}
\end{minipage}
\begin{minipage}{.15\linewidth}
    \centering
    \includegraphics[width=\linewidth]{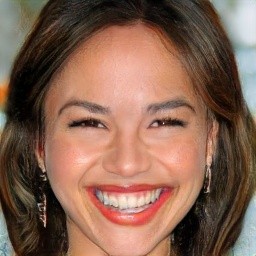}
\end{minipage}
\begin{minipage}{.15\linewidth}
    \centering
    \includegraphics[width=\linewidth]{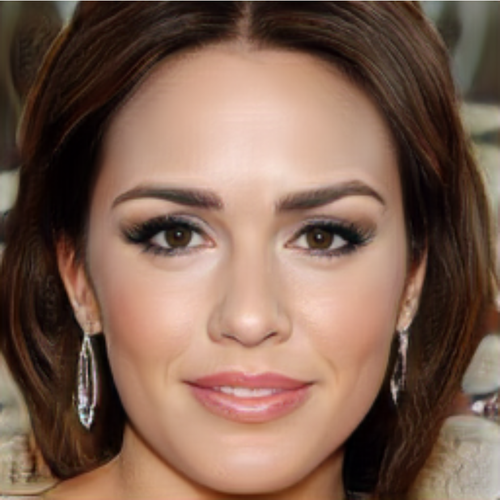}
\end{minipage}
\begin{minipage}{.15\linewidth}
    \centering
    \includegraphics[width=\linewidth]{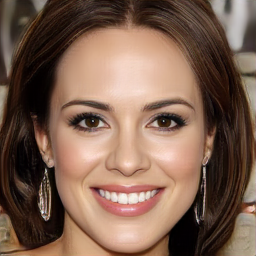}
\end{minipage}
\rotatebox[origin=lc]{90}{\centering \hspace{-0.63cm} \scriptsize{\textbf{Smiling (+)}}}

\rotatebox[origin=lc]{90}{\centering \hspace{-0.45cm} \scriptsize{\textbf{EG3D}}}
\begin{minipage}{.15\linewidth}
    \centering
    \includegraphics[width=\linewidth]{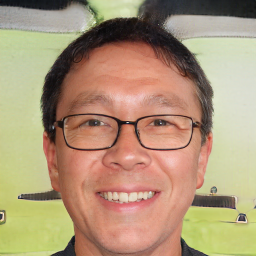}
\end{minipage}
\begin{minipage}{.15\linewidth}
    \centering
    \includegraphics[width=\linewidth]{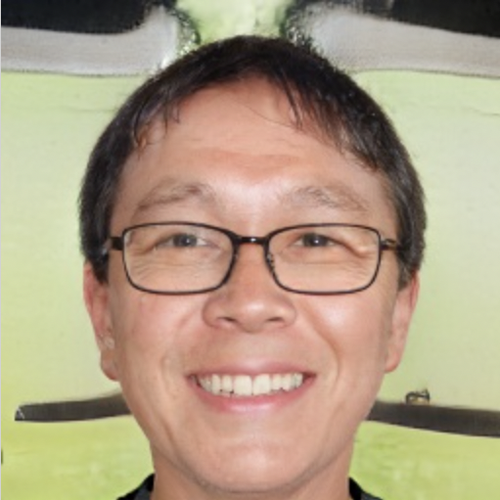}
\end{minipage}
\begin{minipage}{.15\linewidth}
    \centering
    \includegraphics[width=\linewidth]{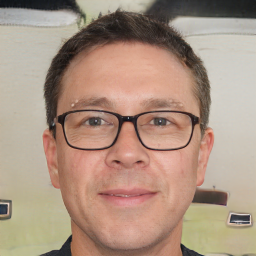}
\end{minipage}
\begin{minipage}{.15\linewidth}
    \centering
    \includegraphics[width=\linewidth]{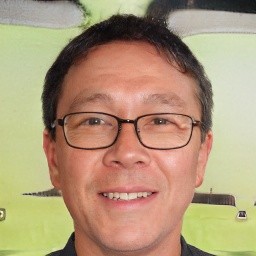}
\end{minipage}
\begin{minipage}{.15\linewidth}
    \centering
    \includegraphics[width=\linewidth]{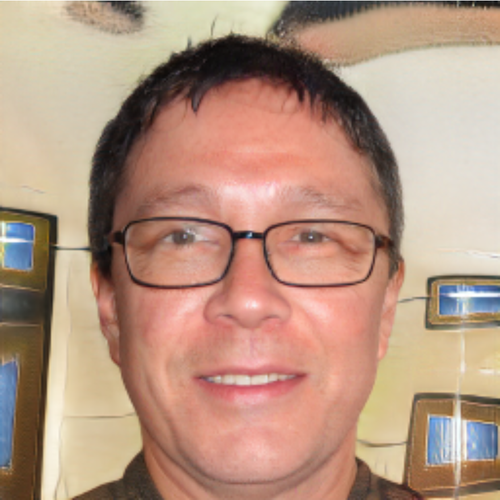}
\end{minipage}
\begin{minipage}{.15\linewidth}
    \centering
    \includegraphics[width=\linewidth]{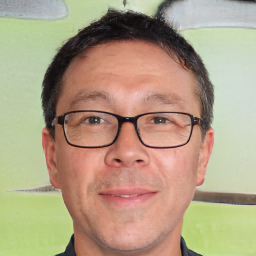}
\end{minipage}
\rotatebox[origin=lc]{90}{\centering \hspace{-0.63cm} \scriptsize{\textbf{Smiling (-)}}}

\caption{Comparison between InterFaceGAN~\cite{shen2020interfacegan}, SeFa~\cite{shen2021closed}, LatentCLR~\cite{yuksel2021latentclr}, StyleFlow~\cite{abdal2021styleflow} and \model{} on \pigan{}, MVCGAN, and EG3D models while performing a \textit{smiling} edit.
\label{fig:comparison_pigan}}
\end{figure}

\noindent\textbf{2D editing methods on 3D GANs.} 
We compare our method with the state-of-the-art 2D-based latent space manipulators, namely, InterFaceGAN~\cite{shen2020interfacegan}, SeFa~\cite{yuksel2021latentclr}, LatentCLR~\cite{shen2021closed}, and StyleFlow~\cite{abdal2021styleflow}. In Fig.~\ref{fig:comparison_pigan}, we show a smile edit on \pigan{}, MVCGAN, and EG3D. 
For SeFa and LatentCLR, identified directions that roughly correspond to the desired edits have been manually selected. Our method provides impressive results for \pigan{}, MVCGAN, and EG3D generators, whereas the other methods sometimes result in nonsensical images or entangled edits. Since InterFaceGAN applies linear operations, if the coefficients of the manipulation are too large, they might conflict with the periodicity of latent space and generate unnatural images (such as the face edit on \pigan{}). StyleFlow changes the identity and fails to apply the desired attributes. On the other hand, SeFa and LatentCLR provide unsupervised edits, but there are no semantically meaningful edits, and sometimes they cannot preserve the identity.

\begin{figure}[ht!]

\centering 
\scriptsize

\rotatebox[origin=lc]{90}{\centering \hspace{-0.45cm} \textbf{\phantom{a}}}
\begin{minipage}{.15\linewidth}
    \centering
    \textbf{Input \\Image}
\end{minipage}
\begin{minipage}{.15\linewidth}
    \centering
    \textbf{Top-128 \\(7\%)}
\end{minipage}
\begin{minipage}{.15\linewidth}
    \centering
    \textbf{Top-256 \\(12\%)}
\end{minipage}
\begin{minipage}{.15\linewidth}
    \centering
    \textbf{Top-512 \\(18\%)}
\end{minipage}
\begin{minipage}{.15\linewidth}
    \centering
    \textbf{Top-1024 \\(36\%)}
\end{minipage}
\begin{minipage}{.15\linewidth}
    \centering
    \textbf{Top-2048 \\(46\%)}
\end{minipage}

\rotatebox[origin=lc]{90}{\centering \hspace{-0.45cm} \textbf{Blonde}}
\begin{minipage}{.15\linewidth}
    \centering
    \includegraphics[width=\linewidth]{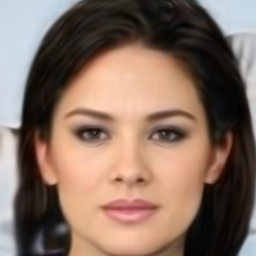}
\end{minipage}
\begin{minipage}{.15\linewidth}
    \centering
    \includegraphics[width=\linewidth]{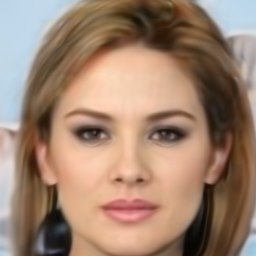}
\end{minipage}
\begin{minipage}{.15\linewidth}
    \centering
    \includegraphics[width=\linewidth]{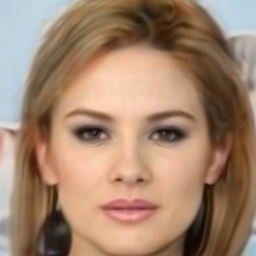}
\end{minipage}
\begin{minipage}{.15\linewidth}
    \centering
    \textcolor{red}{\frame{\includegraphics[width=\linewidth]{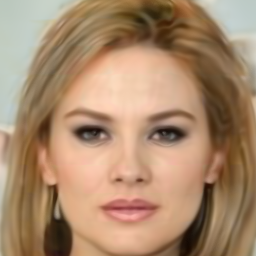}}}
\end{minipage}
\begin{minipage}{.15\linewidth}
    \centering
    \includegraphics[width=\linewidth]{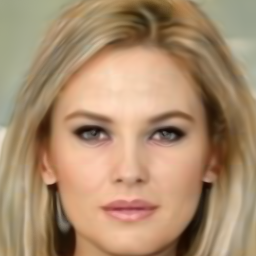}
\end{minipage}
\begin{minipage}{.15\linewidth}
    \centering
    \includegraphics[width=\linewidth]{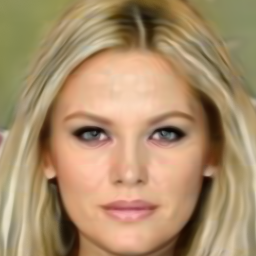}
\end{minipage}

\rotatebox[origin=lc]{90}{\centering \hspace{-0.45cm} \textbf{Gender}}
\begin{minipage}{.15\linewidth}
    \centering
    \includegraphics[width=\linewidth]{figures/pigan/topk/2107665657_original.png}
\end{minipage}
\begin{minipage}{.15\linewidth}
    \centering
    \includegraphics[width=\linewidth]{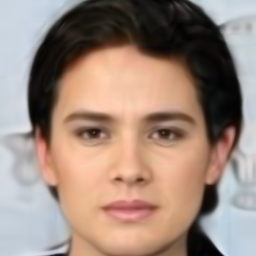}
\end{minipage}
\begin{minipage}{.15\linewidth}
    \centering
    \textcolor{red}{\frame{\includegraphics[width=\linewidth]{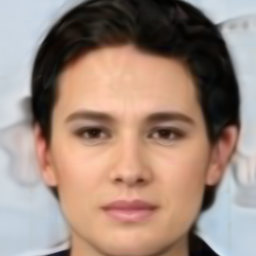}}}
\end{minipage}
\begin{minipage}{.15\linewidth}
    \centering
    \includegraphics[width=\linewidth]{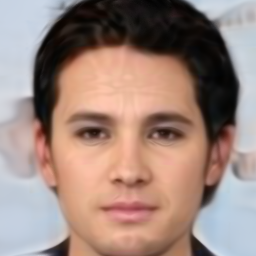}
\end{minipage}
\begin{minipage}{.15\linewidth}
    \centering
    \includegraphics[width=\linewidth]{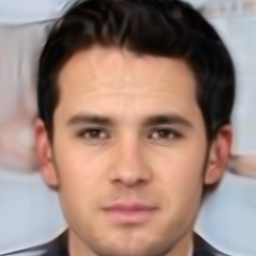}
\end{minipage}
\begin{minipage}{.15\linewidth}
    \centering
    \includegraphics[width=\linewidth]{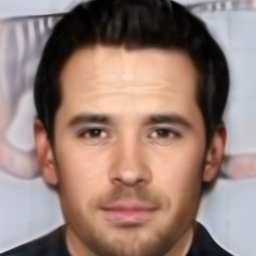}
\end{minipage}

\rotatebox[origin=lc]{90}{\centering \hspace{-0.45cm} \textbf{\phantom{a}}}
\begin{minipage}{.15\linewidth}
    \centering
    \phantom{a}
\end{minipage}
\begin{minipage}{.15\linewidth}
    \centering
    \textbf{(13\%)}
\end{minipage}
\begin{minipage}{.15\linewidth}
    \centering
    \textbf{(24\%)}
\end{minipage}
\begin{minipage}{.15\linewidth}
    \centering
    \textbf{(60\%)}
\end{minipage}
\begin{minipage}{.15\linewidth}
    \centering
    \textbf{(85\%)}
\end{minipage}
\begin{minipage}{.15\linewidth}
    \centering
    \textbf{(96\%)}
\end{minipage}

\caption{The \% corresponds to the identity loss $\mathcal{L}_{ID}$ between edited and original as described in Sec.~\ref{sec:editing}}
\label{fig:topk}

\end{figure}

\noindent\textbf{Impact of parameter top-$K$.} %
Figure~\ref{fig:topk} shows a qualitative example that emphasizes the impact of the number of dimension $K$ swapped for two attribute edits. Higher $K$ values increase the strength of the edit but simultaneously result in images less similar to the input. 
The percentage reported above/below each sample shows the value of identity loss $\mathcal{L}_{ID}$, in Fig.~\ref{fig:topk}. Increasing $K$ results in images less similar to the input image but more similar to the reference image.
We automatically set $K$ for each sample being edited such that $\mathcal{L}_{ID}$, corresponds to $\tau$, does not go above  25\% for face datasets and 10\% for other domains (animals). 
The rightmost faces are the most similar in the support set.

\begin{figure}[ht!]

\centering 

\vspace{0.3cm}

\scriptsize

\rotatebox[origin=lc]{90}{\centering \hspace{-0.45cm} \textbf{ \phantom{Fy}}}
\begin{minipage}{.18\linewidth}
    \centering
    \textbf{Input}
\end{minipage}
\begin{minipage}{.18\linewidth}
    \centering
    \textbf{Eyeglasses (+)}
\end{minipage}
\begin{minipage}{.18\linewidth}
    \centering
    \textbf{Smiling (-)}
\end{minipage}
\begin{minipage}{.18\linewidth}
    \centering
    \textbf{Age (+)}
\end{minipage}
\begin{minipage}{.18\linewidth}
    \centering
    \textbf{Blond Hair}
\end{minipage}

\vspace{0.1cm}

\rotatebox[origin=lc]{90}{\centering \hspace{-0.45cm} \textbf{FFHQ \phantom{y}}}
\begin{minipage}{.18\linewidth}
    \centering
    \includegraphics[width=\linewidth]{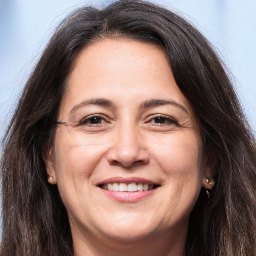}
\end{minipage}
\begin{minipage}{.18\linewidth}
    \centering
    \includegraphics[width=\linewidth]{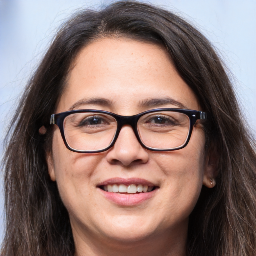}
\end{minipage}
\begin{minipage}{.18\linewidth}
    \centering
    \includegraphics[width=\linewidth]{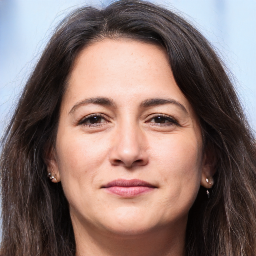}
\end{minipage}
\begin{minipage}{.18\linewidth}
    \centering
    \includegraphics[width=\linewidth]{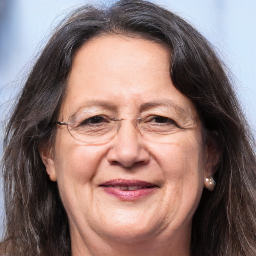}
\end{minipage}
\begin{minipage}{.18\linewidth}
    \centering
    \includegraphics[width=\linewidth]{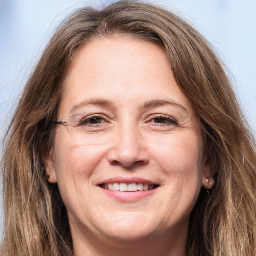}
\end{minipage}

\vspace{0.1cm}

\rotatebox[origin=lc]{90}{\centering \hspace{-0.45cm} \textbf{ \phantom{Fy}}}
\begin{minipage}{.18\linewidth}
    \centering
    \textbf{Input}
\end{minipage}
\begin{minipage}{.18\linewidth}
    \centering
    \textbf{Age (+)}
\end{minipage}
\begin{minipage}{.18\linewidth}
    \centering
    \textbf{Black Hair}
\end{minipage}
\begin{minipage}{.18\linewidth}
    \centering
    \textbf{Brown Hair}
\end{minipage}
\begin{minipage}{.18\linewidth}
    \centering
    \textbf{Gray Hair}
\end{minipage}

\vspace{0.1cm}

\rotatebox[origin=lc]{90}{\centering \hspace{-0.55cm} \textbf{MetFaces \phantom{y}}}
\begin{minipage}{.18\linewidth}
    \centering
    \includegraphics[width=\linewidth]{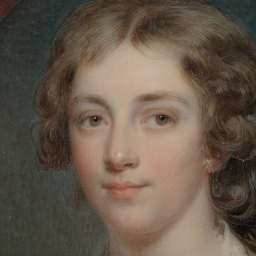}
\end{minipage}
\begin{minipage}{.18\linewidth}
    \centering
    \includegraphics[width=\linewidth]{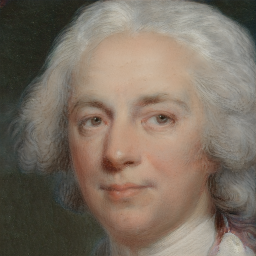}
\end{minipage}
\begin{minipage}{.18\linewidth}
    \centering
    \includegraphics[width=\linewidth]{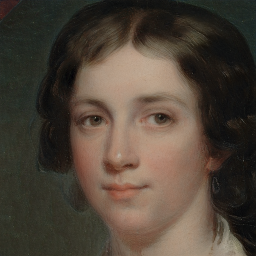}
\end{minipage}
\begin{minipage}{.18\linewidth}
    \centering
    \includegraphics[width=\linewidth]{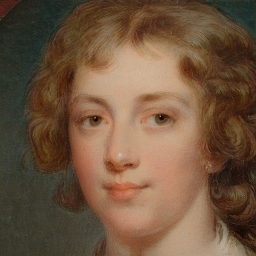}
\end{minipage}
\begin{minipage}{.18\linewidth}
    \centering
    \includegraphics[width=\linewidth]{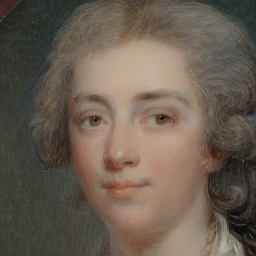}
\end{minipage}

\vspace{0.05cm}

\rotatebox[origin=lc]{90}{\centering \hspace{-0.45cm} \textbf{ \phantom{Fy}}}
\begin{minipage}{.18\linewidth}
    \centering
    \textbf{Input}
\end{minipage}
\begin{minipage}{.18\linewidth}
    \centering
    \textbf{Breed}
\end{minipage}
\begin{minipage}{.18\linewidth}
    \centering
    \textbf{Input}
\end{minipage}
\begin{minipage}{.18\linewidth}
    \centering
    \textbf{Breed}
\end{minipage}
\begin{minipage}{.18\linewidth}
    \centering
    \textbf{Fur Color}
\end{minipage}

\vspace{-0.05cm}

\rotatebox[origin=lc]{90}{\centering \hspace{-0.65cm} \textbf{AFHQ-Cats \phantom{y}}}
\begin{minipage}{.18\linewidth}
    \centering
    \includegraphics[width=\linewidth]{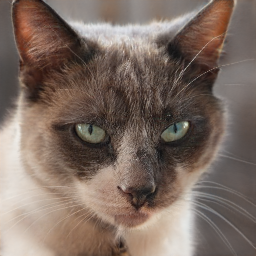}
\end{minipage}
\begin{minipage}{.18\linewidth}
    \centering
    \includegraphics[width=\linewidth]{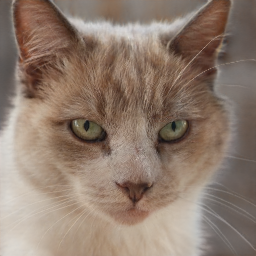}
\end{minipage}
\begin{minipage}{.18\linewidth}
    \centering
    \includegraphics[width=\linewidth]{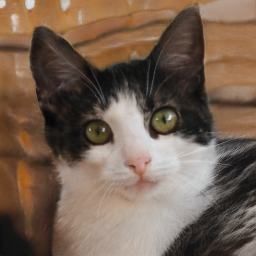}
\end{minipage}
\begin{minipage}{.18\linewidth}
    \centering
    \includegraphics[width=\linewidth]{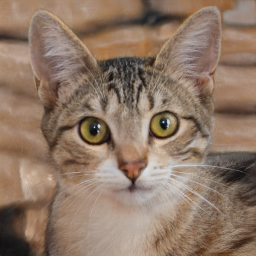}
\end{minipage}
\begin{minipage}{.18\linewidth}
    \centering
    \includegraphics[width=\linewidth]{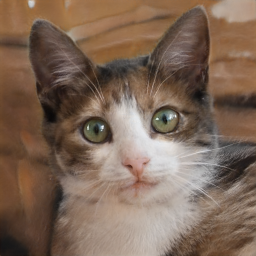}
\end{minipage}

\caption{Several attribute edits for human and animal faces on StyleGAN2~\cite{Karras2020stylegan2} for AFHQ~\cite{choi2020starganv2}, MetFaces~\cite{Karras2020ada}, and FFHQ~\cite{karras2019style} by using \model{}.}
\label{fig:results_stylegan2}
\end{figure}

\noindent\textbf{Proposed method on StyleGAN2.} 
\model{} is not limited to 3D GANs but also works \emph{without modifications} on image-based GANs like StyleGAN2, see Fig.~\ref{fig:results_stylegan2}. First, by applying the procedure in Sec.~\ref{sec:ranking}, we identify the latent codes from the style space of StyleGAN2 that are most important for the desired attribute. Then, we swap those latent codes to generate the desired edits, as explained in Sec.~\ref{sec:editing}. 

\subsection{Quantitative Evaluation} \label{sec:quantitative}

\noindent\textbf{Semantic Correctness.} To evaluate the effectiveness of our attribute edits quantitatively, we use a pre-trained smile attribute classifier~\cite{serengil2021lightface} to measure the percentage of smiling images in a set of 2500 images of not smiling faces. We edit the images with different methods and measure the increase in percentage. For the result, see Tab.~\ref{tab:semantic_correctness}. With \pigan{}, our method increases the percentage of smiling images by $84\%$ whereas InterFaceGAN and StyleFlow increase only by $77\%$ and $79\%$, respectively. The same improvement applies to images generated by MVCGAN and EG3D. Percentages of improvement for each of them are $92\%$ and $84\%$, respectively. 

{\small
\begin{table}[!ht]

\begin{center}
\begin{tabular}{|l|c|c|c|}
\hline
&  \pigan{} &  MVCGAN & EG3D \\
\hline

Unedited Images & 4\% & 3\% & 9\% \\ 
\hline

InterFaceGAN~\cite{shen2020interfacegan} & 81\% & 84\% & 85\% \\ 
\hline
StyleFlow~\cite{abdal2021styleflow} & 83\% & 78\% & 88\% \\ 
\hline
Ours~(\model{}) &  \textbf{88\%} & \textbf{95\%} & \textbf{93\%} \\ 
\hline
\end{tabular}  

\end{center}
\vspace{-0.4cm}

\caption{Semantic correctness metric among different image editing methods for \pigan{}~\cite{pigan}, MVCGAN~\cite{zhang2022multi}, and EG3D\cite{chan2021efficient} on smiling attribute edits of face images.
\label{tab:semantic_correctness} } 
\vspace{-0.1cm}

\end{table}
}

\noindent\textbf{Identity preservation.} We measure the identity preservation between input and edited images using an identity verification tool~\cite{serengil2021lightface} based on FaceNet512~\cite{schroff2015facenet}. Table~\ref{tab:identity_preservation} shows the identity preservation metric on 10K images, and compared to the other methods, our method is the best to preserve the identity of the input image.

{\small
\begin{table}[!ht]

\begin{center}
\begin{tabular}{|l|c|c|c|}
\hline
&  \pigan{} &  MVCGAN & EG3D \\
\hline

LatentCLR~\cite{shen2021closed} & 54\% & 61\% & 69\% \\ 
\hline
SeFa~\cite{yuksel2021latentclr} & 62\% & 64\% & 58\% \\ 
\hline
InterFaceGAN~\cite{shen2020interfacegan} & 30\% & 51\% & 71\% \\ 
\hline
StyleFlow~\cite{abdal2021styleflow} & 68\% & 65\% & 72\% \\ 
\hline
Ours~(\model{}) &  \textbf{74\%} & \textbf{71\%} & \textbf{73\%} \\ 
\hline
\end{tabular}  

\end{center}
\vspace{-0.4cm}

\caption{Identity preservation metric among different image editing methods for \pigan{}~\cite{pigan}, MVCGAN~\cite{zhang2022multi}, and EG3D\cite{chan2021efficient} on several attribute edits of face images.
\label{tab:identity_preservation} } 
\vspace{-0.1cm}

\end{table}
}

\section{Conclusions}
\label{sec:conclusions}
To the best of our knowledge, we propose the first generator- and dataset-agnostic  semantic editing method for 3D GANs. We show this by applying our method to various generators (\eg{,} \pigan{}, GIRAFFE, StyleSDF, MVCGAN, EG3D and VolumeGAN) and datasets (\eg{}, FFHQ, AFHQ, Cats, MetFaces, and CompCars). Additionally, our method enables complex edits and multi-view consistent rendering from a single image of a real face or an object, opening the path to multiple practical applications.
The broader impact of this work includes possible use cases in compression for video conferencing or 3D manipulation for AR overlays. On the other hand, like all GAN-based image editing methods, \model{} will suffer from datasets bias and is limited by the images that can be modeled by the GAN being manipulated.
However, considering the rapid progress in generative modeling and the generality of our proposed framework, we envision that our method will be equally applicable in future generations of generative models, resulting in even more impressive editing capabilities.

\paragraph{\textbf{Acknowledgements}}
We are grateful to Google University Relationship GCP Credit Program for the support of this work by providing computational resources.

{
    \small
    \bibliographystyle{ieeenat_fullname}
    \bibliography{egbib}
}

\clearpage
\setcounter{page}{1}

\twocolumn[{%
\renewcommand\twocolumn[1][]{#1}%

\maketitlesupplementary

\begin{center}
\centering

\small

\begin{minipage}{.30\linewidth}
    \centering
    \textbf{\pigan{}}
\end{minipage}
\begin{minipage}{.30\linewidth}
    \centering
    \textbf{MVCGAN}
\end{minipage}
\begin{minipage}{.30\linewidth}
    \centering
    \textbf{EG3D}
\end{minipage}
\rotatebox[origin=lc]{90}{\centering \small{\phantom{aa}}}

\begin{minipage}{.10\linewidth}
    \centering
    \textbf{Input}
\end{minipage}
\begin{minipage}{.10\linewidth}
    \centering
    \textbf{Gender}
\end{minipage}
\begin{minipage}{.10\linewidth}
    \centering
    \textbf{Black Hair}
\end{minipage}
\begin{minipage}{.10\linewidth}
    \centering
    \textbf{Input}
\end{minipage}
\begin{minipage}{.10\linewidth}
    \centering
    \textbf{Bangs}
\end{minipage}
\begin{minipage}{.10\linewidth}
    \centering
    \textbf{Blond Hair}
\end{minipage}
\begin{minipage}{.10\linewidth}
    \centering
    \textbf{Input}
\end{minipage}
\begin{minipage}{.10\linewidth}
    \centering
    \textbf{Bangs}
\end{minipage}
\begin{minipage}{.10\linewidth}
    \centering
    \textbf{Age}
\end{minipage}
\rotatebox[origin=lc]{90}{\centering \small{\phantom{aa}}}

\begin{minipage}{.10\linewidth}
    \centering
    \includegraphics[width=\linewidth]{figures/comparison/pigan/69_original.png}
\end{minipage}
\begin{minipage}{.10\linewidth}
    \centering
    \includegraphics[width=\linewidth]{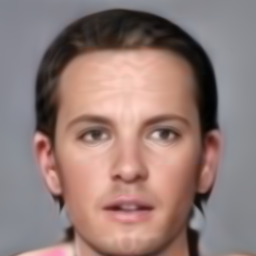}
\end{minipage}
\begin{minipage}{.10\linewidth}
    \centering
    \includegraphics[width=\linewidth]{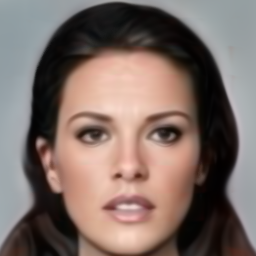}
\end{minipage}
\begin{minipage}{.10\linewidth}
    \centering
    \includegraphics[width=\linewidth]{figures/comparison/mvcgan/2_original.png}
\end{minipage}
\begin{minipage}{.10\linewidth}
    \centering
    \includegraphics[width=\linewidth]{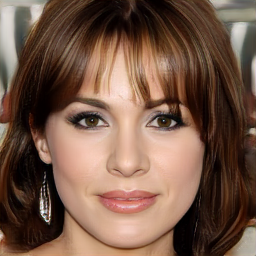}
\end{minipage}
\begin{minipage}{.10\linewidth}
    \centering
    \includegraphics[width=\linewidth]{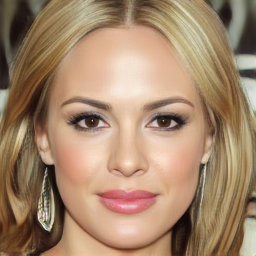}
\end{minipage}
\begin{minipage}{.10\linewidth}
    \centering
    \includegraphics[width=\linewidth]{figures/comparison/eg3d/1_original.png}
\end{minipage}
\begin{minipage}{.10\linewidth}
    \centering
    \includegraphics[width=\linewidth]{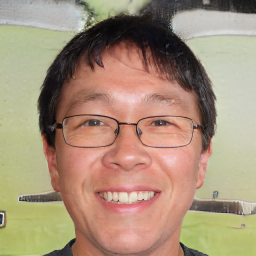}
\end{minipage}
\begin{minipage}{.10\linewidth}
    \centering
    \includegraphics[width=\linewidth]{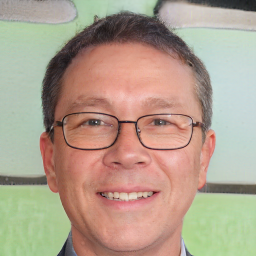}
\end{minipage}
\begin{minipage}{.02\linewidth}
    \raggedleft
    \rotatebox[origin=lc]{90}{\centering \small{\textbf{Ours}}}
\end{minipage}

\begin{minipage}{.10\linewidth}
    \centering
    \phantom{\includegraphics[width=\linewidth]{figures/pigan/topk/2107665657_original.png}}
\end{minipage}
\begin{minipage}{.10\linewidth}
    \centering
    \includegraphics[width=\linewidth]{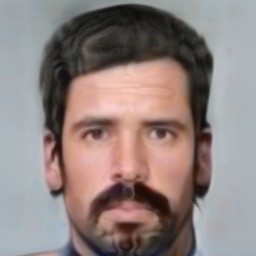}
\end{minipage}
\begin{minipage}{.10\linewidth}
    \centering
    \includegraphics[width=\linewidth]{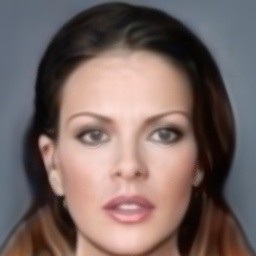}
\end{minipage}
\begin{minipage}{.10\linewidth}
    \centering
    \phantom{\includegraphics[width=\linewidth]{figures/comparison/mvcgan/2_original.png}}
\end{minipage}
\begin{minipage}{.10\linewidth}
    \centering
    \includegraphics[width=\linewidth]{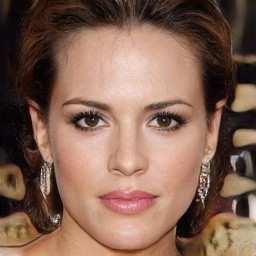}
\end{minipage}
\begin{minipage}{.10\linewidth}
    \centering
    \includegraphics[width=\linewidth]{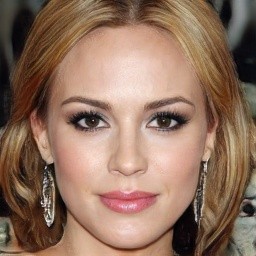}
\end{minipage}
\begin{minipage}{.10\linewidth}
    \centering
    \phantom{\includegraphics[width=\linewidth]{figures/comparison/mvcgan/2_original.png}}
\end{minipage}
\begin{minipage}{.10\linewidth}
    \centering
    \includegraphics[width=\linewidth]{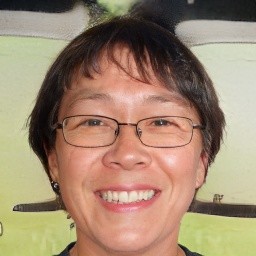}
\end{minipage}
\begin{minipage}{.10\linewidth}
    \centering
    \includegraphics[width=\linewidth]{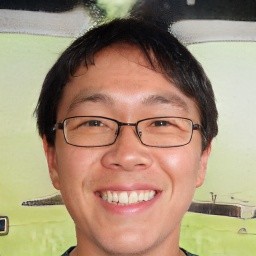}
\end{minipage}
\begin{minipage}{.02\linewidth}
    \raggedleft
    \rotatebox[origin=lc]{90}{\centering \small{\textbf{IGAN~\cite{shen2020interfacegan}}}}
\end{minipage}

\begin{minipage}{.10\linewidth}
    \centering
    \phantom{\includegraphics[width=\linewidth]{figures/pigan/topk/2107665657_original.png}}
\end{minipage}
\begin{minipage}{.10\linewidth}
    \centering
    \includegraphics[width=\linewidth]{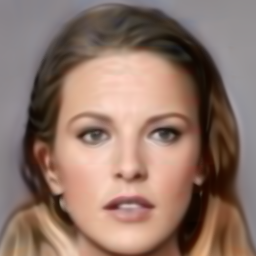}
\end{minipage}
\begin{minipage}{.10\linewidth}
    \centering
    \includegraphics[width=\linewidth]{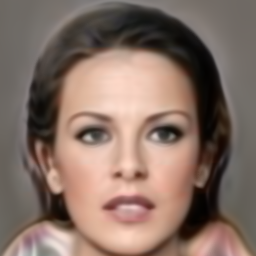}
\end{minipage}
\begin{minipage}{.10\linewidth}
    \centering
    \phantom{\includegraphics[width=\linewidth]{figures/pigan/topk/2107665657_original.png}}
\end{minipage}
\begin{minipage}{.10\linewidth}
    \centering
    \includegraphics[width=\linewidth]{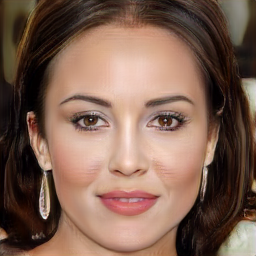}
\end{minipage}
\begin{minipage}{.10\linewidth}
    \centering
    \includegraphics[width=\linewidth]{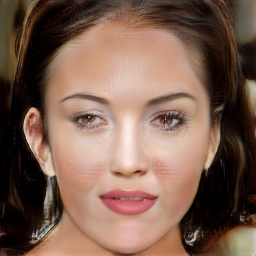}
\end{minipage}
\begin{minipage}{.10\linewidth}
    \centering
    \phantom{\includegraphics[width=\linewidth]{figures/pigan/topk/2107665657_original.png}}
\end{minipage}
\begin{minipage}{.10\linewidth}
    \centering
    \includegraphics[width=\linewidth]{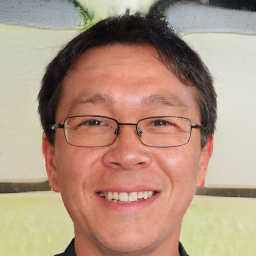}
\end{minipage}
\begin{minipage}{.10\linewidth}
    \centering
    \includegraphics[width=\linewidth]{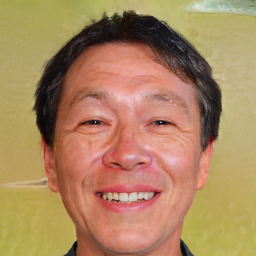}
\end{minipage}
\begin{minipage}{.02\linewidth}
    \raggedleft
    \rotatebox[origin=lc]{90}{\centering \small{\textbf{SFlow.~\cite{abdal2021styleflow}}}}
\end{minipage}

\captionof{figure}{Detailed comparison to other methods for \pigan{}, MVCGAN and EG3D. The edited attributes are annotated. \label{fig:comparison_all}}
\vspace{0.4cm}

\end{center}

}]

We first provide additional quantitative and qualitative results for our \model{}. Then, we show ablation studies on hyper-parameters, alternative edit techniques and feature ranking methods. Finally, we discuss limitations, implementation details, and future work.

\section{Additional Results}\label{sec:additional_qualitative}
\subsection{Additional Comparison to Other Methods} 

In addition to Fig.~\ref{fig:comparison_pigan}, we report here an additional comparison among \model{}, InterFaceGAN~\cite{shen2020interfacegan} and StyleFlow~\cite{abdal2021styleflow}.
As seen in~\cref{fig:comparison_all}, our approach provides meaningful semantic edits without changing the identity on the input image and performs best for all desired attributes among 3D GANs, while the other methods may change the identity or make entangled edits.

\begin{figure}[ht!]
\centering

\small
\begin{minipage}{.05\linewidth}
    \raggedleft
    \rotatebox[origin=lc]{90}{\centering \phantom{\small{\textbf{k}}}}
\end{minipage}
\begin{minipage}{.18\linewidth}
    \centering
    \textbf{Input}
\end{minipage}
\begin{minipage}{.54\linewidth}
    \centering
    \textbf{Breed}
\end{minipage}

\begin{minipage}{.05\linewidth}
    \raggedleft
    \rotatebox[origin=lc]{90}{\centering \small{\textbf{\pigan{}}}}
\end{minipage}
\begin{minipage}{.18\linewidth}
    \centering
    \includegraphics[width=\linewidth]{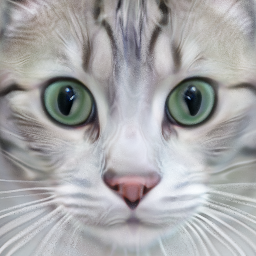}
\end{minipage}
\begin{minipage}{.54\linewidth}
    \centering
    \includegraphics[width=\linewidth]{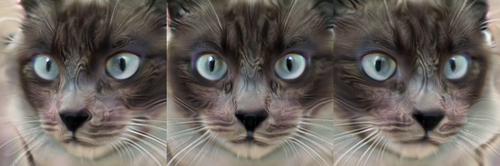}
\end{minipage}

\begin{minipage}{.05\linewidth}
    \raggedleft
    \rotatebox[origin=lc]{90}{\centering \small{\textbf{MVCGAN}}}
\end{minipage}
\begin{minipage}{.18\linewidth}
    \centering
    \includegraphics[width=\linewidth]{figures/mvcgan/cats/1360262803_original.png}
\end{minipage}
\begin{minipage}{.54\linewidth}
    \centering
    \includegraphics[width=\linewidth]{figures/mvcgan/cats/1360262803_siamese.png}
\end{minipage}

\vspace{0.03cm}

\rotatebox[origin=lc]{90}{\centering \hspace{-0.35cm}
\phantom{\small{\textbf{k}}}}
\begin{minipage}{.85\linewidth}
    \centering
    \hrule width\linewidth\relax
\end{minipage}

\vspace{-0.03cm}

\begin{minipage}{.05\linewidth}
    \raggedleft
    \rotatebox[origin=lc]{90}{\centering \small{\textbf{\pigan{}}}}
\end{minipage}
\begin{minipage}{.18\linewidth}
    \centering
    \includegraphics[width=\linewidth]{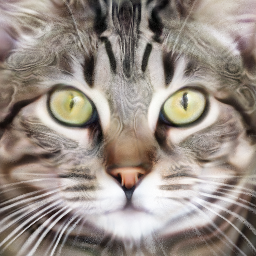}
\end{minipage}
\begin{minipage}{.54\linewidth}
    \centering
    \includegraphics[width=\linewidth]{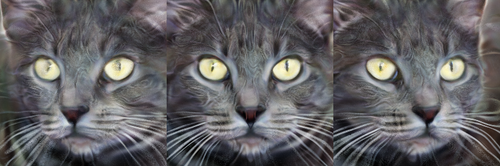}
\end{minipage}

\begin{minipage}{.05\linewidth}
    \raggedleft
    \rotatebox[origin=lc]{90}{\centering \small{\textbf{EG3D}}}
\end{minipage}
\begin{minipage}{.18\linewidth}
    \centering
    \includegraphics[width=\linewidth]{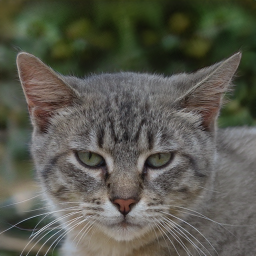}
\end{minipage}
\begin{minipage}{.54\linewidth}
    \centering
    \includegraphics[width=\linewidth]{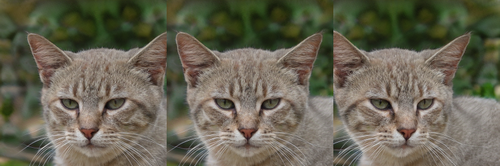}
\end{minipage}

\rotatebox[origin=lc]{90}{\centering \hspace{-0.35cm}
\phantom{\small{\textbf{k}}}}
\begin{minipage}{.18\linewidth}
    \centering
    \textbf{Input}
\end{minipage}
\begin{minipage}{.54\linewidth}
    \centering
    \textbf{Color}
\end{minipage}
\vspace{-0.1cm}

\caption{Results for Cats dataset~\cite{cats} with \pigan{} and AFHQ dataset~\cite{choi2020starganv2} with MVCGAN and EG3D generators. 
}
\label{fig:additional_cat_examples}
\vspace{-0.3cm}
\end{figure}

\subsection{Additional Animal Editing} 
To further support our findings, we included additional attribute editing examples for animals in addition to Fig.~\ref{fig:results_cats}. Figure~\ref{fig:additional_cat_examples} shows the successful edits of color and breed attributes on pre-trained \pigan{}, MVCGAN, and EG3D generators using our proposed \model{}.

\begin{figure}[ht!]
\vspace{-0.1cm}
\centering 
\scriptsize

\rotatebox[origin=lc]{90}{\centering \hspace{-0.5cm} \textbf{ \phantom{Fy}}}
\begin{minipage}{.18\linewidth}
    \centering
    \textbf{\scriptsize{Input}}
\end{minipage}
\begin{minipage}{.18\linewidth}
    \centering
    \textbf{\scriptsize{Breed Change}}
\end{minipage}
\begin{minipage}{.02\linewidth}
    \centering
    \phantom{\includegraphics[width=\linewidth]{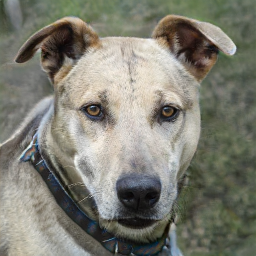}}
\end{minipage}
\begin{minipage}{.18\linewidth}
    \centering
    \textbf{\scriptsize{Input}}
\end{minipage}
\begin{minipage}{.36\linewidth}
    \centering
    \textbf{\scriptsize{Fur Color Change}}
\end{minipage}

\begin{minipage}{.03\linewidth}
    \raggedleft
    \rotatebox[origin=lc]{90}{\centering \textbf{AFHQ-Dogs\phantom{y}}}
\end{minipage}
\begin{minipage}{.18\linewidth}
    \centering
    \includegraphics[width=\linewidth]{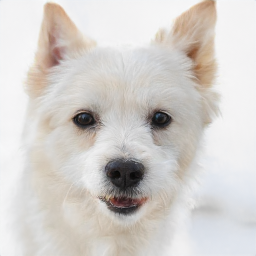}
\end{minipage}
\begin{minipage}{.18\linewidth}
    \centering
    \includegraphics[width=\linewidth]{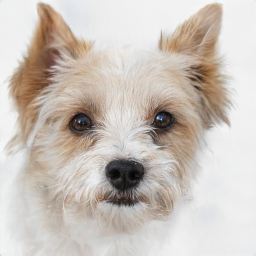}
\end{minipage}
\begin{minipage}{.02\linewidth}
    \centering
    \phantom{\includegraphics[width=\linewidth]{figures/stylegan2/dogs/1915217934_original.png}}
\end{minipage}
\begin{minipage}{.18\linewidth}
    \centering
    \includegraphics[width=\linewidth]{figures/stylegan2/dogs/1915217934_original.png}
\end{minipage}
\begin{minipage}{.18\linewidth}
    \centering
    \includegraphics[width=\linewidth]{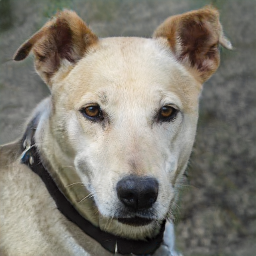}
\end{minipage}
\begin{minipage}{.18\linewidth}
    \centering
    \includegraphics[width=\linewidth]{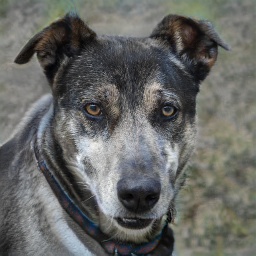}
\end{minipage}

\caption{Results for AFHQ dataset~\cite{choi2020starganv2}, for dog images, from \model{} on StyleGAN2.}
\label{fig:results_stylegan2_dogs}
\vspace{-0.2cm}
\end{figure}

In \cref{fig:results_stylegan2_dogs}, we show additional qualitative results on applying edits to samples generated from a StyleGAN model trained on AFHQ~\cite{choi2020starganv2} - Dogs dataset. We use the attribute classifiers presented in \cref{sec:animal_classifier} to identify the dimension to edit. Note that even if the classifiers are trained on cat images, they can successfully be used for attribute editing on dog images.

\subsection{Additional Real Image Editing Results}
\begin{figure}[!ht]
    \centering
    
    \begin{minipage}{.2\linewidth}
        \centering
        \textbf{Input}
    \end{minipage}
    \begin{minipage}{.04\linewidth}
        \centering
        \phantom{\textbf{y}}
    \end{minipage}
    \begin{minipage}{.6\linewidth}
        \centering
        \textbf{\model{} on MVCGAN}
    \end{minipage}
    
    \begin{minipage}{.2\linewidth}
        \centering
        \includegraphics[width=\linewidth]{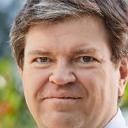}
    \end{minipage}
    \begin{minipage}{.04\linewidth}
        \raggedleft
        \rotatebox[origin=lc]{90}{\centering \hspace{-0.3cm} \small{\textbf{\phantom{y}Inverted}}}
    \end{minipage}
    \begin{minipage}{.6\linewidth}
        \centering
        \includegraphics[width=\linewidth]{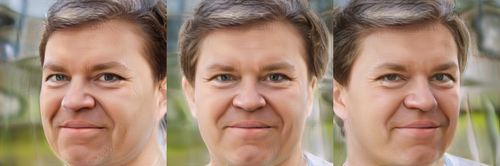}
    \end{minipage}
    
    \begin{minipage}{.2\linewidth}
        \centering
        \phantom{\includegraphics[width=\linewidth]{figures/comparison/confignet/lecun/gt.jpg}}
    \end{minipage}
    \begin{minipage}{.04\linewidth}
        \raggedleft
        \rotatebox[origin=lc]{90}{\centering \hspace{-0.3cm} \small{\textbf{\phantom{y}Smiling (+)}}}
    \end{minipage}
    \begin{minipage}{.6\linewidth}
        \centering
        \includegraphics[width=\linewidth]{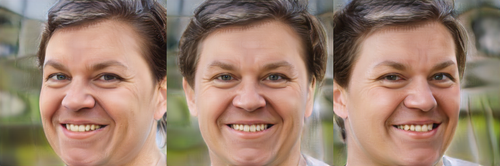}
    \end{minipage}    
    
    \begin{minipage}{.2\linewidth}
        \centering
        \phantom{\includegraphics[width=\linewidth]{figures/comparison/confignet/lecun/gt.jpg}}
    \end{minipage}
    \begin{minipage}{.04\linewidth}
        \raggedleft
        \rotatebox[origin=lc]{90}{\centering \hspace{-0.3cm} \small{\textbf{\phantom{y}Eyeglasses}}}
    \end{minipage}
    \begin{minipage}{.6\linewidth}
        \centering
        \includegraphics[width=\linewidth]{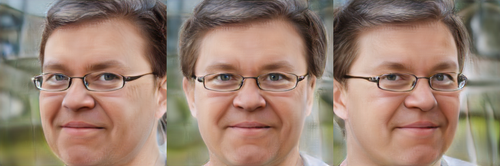}
    \end{minipage}
    
    \begin{minipage}{.2\linewidth}
        \centering
        \includegraphics[width=\linewidth]{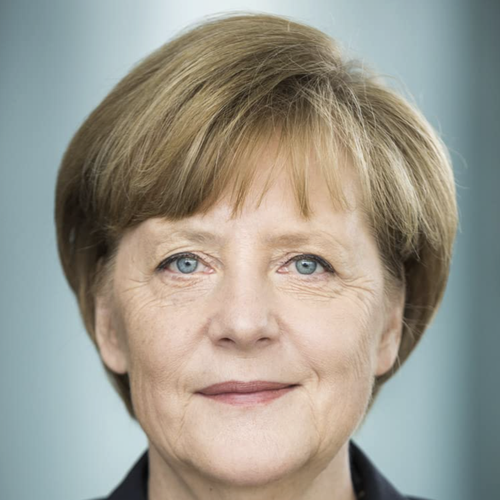}
    \end{minipage}
    \begin{minipage}{.04\linewidth}
        \raggedleft
        \rotatebox[origin=lc]{90}{\centering \hspace{-0.3cm} \small{\textbf{\phantom{y}Inverted}}}
    \end{minipage}
    \begin{minipage}{.6\linewidth}
        \centering
        \includegraphics[width=\linewidth]{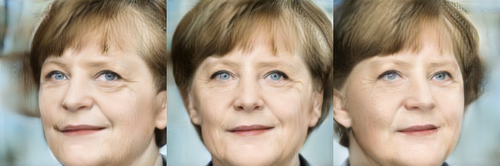}
    \end{minipage}

    \begin{minipage}{.2\linewidth}
        \centering
        \phantom{\includegraphics[width=\linewidth]{figures/comparison/confignet/lecun/gt.jpg}}
    \end{minipage}
    \begin{minipage}{.04\linewidth}
        \raggedleft
        \rotatebox[origin=lc]{90}{\centering \hspace{-0.3cm} \small{\textbf{\phantom{y}Smiling (+)}}}
    \end{minipage}
    \begin{minipage}{.6\linewidth}
        \centering
        \includegraphics[width=\linewidth]{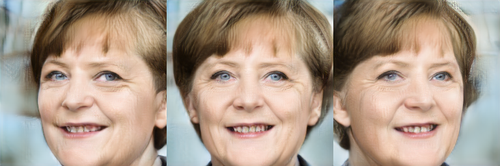}
    \end{minipage}

    \begin{minipage}{.2\linewidth}
        \centering
        \phantom{\includegraphics[width=\linewidth]{figures/comparison/confignet/lecun/gt.jpg}}
    \end{minipage}
    \begin{minipage}{.04\linewidth}
        \raggedleft
        \rotatebox[origin=lc]{90}{\centering \hspace{-0.3cm} \small{\textbf{\phantom{y}Dark Hair}}}
    \end{minipage}
    \begin{minipage}{.6\linewidth}
        \centering
        \includegraphics[width=\linewidth]{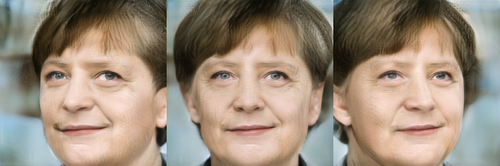  }
    \end{minipage}

    \caption{Additional inverted and edited examples from our approach on MVCGAN.}
    \label{fig:mvcgan_invert}
\end{figure}

We show additional real image editing results for MVCGAN. Thanks to the generator model's high-resolution output, our editing results are also in very high resolution. 
In \cref{fig:mvcgan_invert}, we show results for face inversion and several attribute editings, \eg{,} smiling, changing hair color, and wearing eyeglasses. 
In all cases, our edits correctly maintain the 3D consistency of the generated face.

\subsection{\model{} on Other 3D-aware Generators}

\begin{figure}[!ht]
    \centering
    \small

    \begin{minipage}{.05\linewidth}
        \raggedleft
        \rotatebox[origin=lc]{90}{\centering \small{\textbf{Input}}}
    \end{minipage}
    \begin{minipage}{0.9\linewidth}
        \centering
        \includegraphics[width=\linewidth]{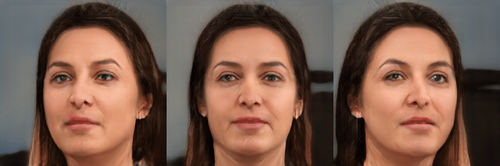}
    \end{minipage}

    \begin{minipage}{.05\linewidth}
        \raggedleft
        \rotatebox[origin=lc]{90}{\centering \small{\textbf{Smiling}}}
    \end{minipage}
    \begin{minipage}{0.9\linewidth}
        \centering
        \includegraphics[width=\linewidth]{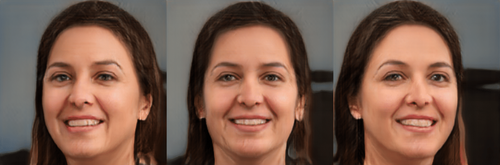}
    \end{minipage}

    \vspace{0.03cm}
    
    \rotatebox[origin=lc]{90}{\centering \hspace{-0.35cm}
    \phantom{\small{\textbf{k}}}}
    \begin{minipage}{.95\linewidth}
        \centering
        \hrule width\linewidth\relax
    \end{minipage}
    
    \vspace{-0.03cm}

    \begin{minipage}{.05\linewidth}
        \raggedleft
        \rotatebox[origin=lc]{90}{\centering \small{\textbf{Input}}}
    \end{minipage}
    \begin{minipage}{0.9\linewidth}
        \centering
        \includegraphics[width=\linewidth]{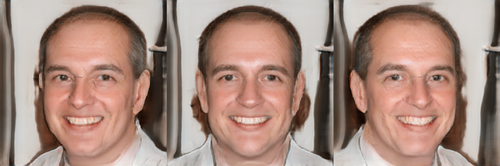}
    \end{minipage}

    \begin{minipage}{.05\linewidth}
        \raggedleft
        \rotatebox[origin=lc]{90}{\centering \small{\textbf{Eyeglasses}}}
    \end{minipage}
    \begin{minipage}{0.9\linewidth}
        \centering
        \includegraphics[width=\linewidth]{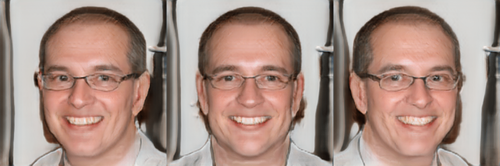}
    \end{minipage}
    
    \caption{LatentSwap3D on GIRAFFE~\cite{Niemeyer2020GIRAFFE} - FFHQ}
    \label{fig:giraffe_ffhq}
\end{figure}

\begin{figure}[!ht]

    \centering

    \begin{minipage}{0.95\linewidth}
        \centering
        \includegraphics[width=\linewidth]{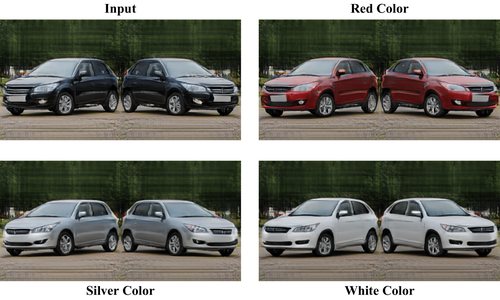}
    \end{minipage}

    \vspace{-0.35cm}
    
    \rotatebox[origin=lc]{90}{\centering \hspace{-0.35cm}
    \phantom{\small{\textbf{k}}}}
    \begin{minipage}{0.98\linewidth}
        \centering
        \hrule width\linewidth\relax
    \end{minipage}
    
    \vspace{-0.03cm}

    \begin{minipage}{0.95\linewidth}
        \centering
        \includegraphics[width=\linewidth]{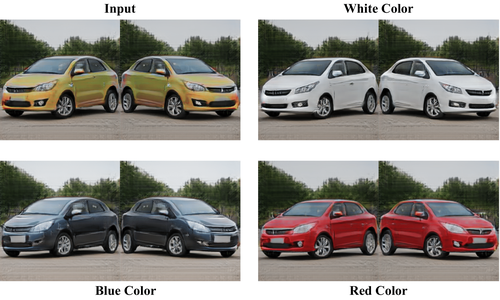}
    \end{minipage}
    
    \caption{LatentSwap3D on GIRAFFE - CompCars~\cite{yang2015large}.}
    \label{fig:giraffe_cars}
\end{figure}

\noindent\textbf{GIRAFFE} consists of NeRF and 2D GANs. The NeRF part outputs the features of the 3D shape and texture, while the 2D GAN part outputs the final image~\cite{Niemeyer2020GIRAFFE}. In \cref{fig:giraffe_ffhq}, we show smiling and wearing eyeglasses edits from \model{} on the GIRAFFE - FFHQ model.
To test how well \model{} generalizes to different datasets, we extended the experiment to include CompCars~\cite{yang2015large} using the pre-trained GIRAFFE generator. Furthermore, due to the lack of classifiers for car attributes, as a proof of concept, we trained a ResNet-50 to classify the color of a car from scratch on Myauto.ge Cars Dataset~\cite{autoge}. As seen from \cref{fig:giraffe_cars}, our approach can successfully edit the color of the cars using these classifiers.

\begin{table*}[!ht]
  \centering
\resizebox{0.9\linewidth}{!}{%
  \begin{tabular}{|c|cc|cc||cc|cc||cc|cc|}
  \hline
    & \multicolumn{4}{c||}{\pigan{}} & \multicolumn{4}{c||}{MVCGAN} & \multicolumn{4}{c|}{EG3D} \\
    \hline
    & \multicolumn{2}{c|}{CelebA} & \multicolumn{2}{c||}{Cats} & \multicolumn{2}{c|}{FFHQ} & \multicolumn{2}{c||}{AFHQ} & \multicolumn{2}{c|}{FFHQ} & \multicolumn{2}{c|}{AFHQ} \\
    Method & FID & KID & FID & KID & FID & KID & FID & KID & FID & KID & FID & KID \\
    \hline 
    Unedited & 50.7 & 0.045 & 57.4 & 0.055 & 54.1 & 0.048 & 47.1 & 0.041 & 47.5 & 0.039 & 39.5 & 0.037 \\
    \hline 
    \hline
    LCLR. & 53.4 & 0.051 & 60.1 & 0.062 & \textbf{55.6} & 0.052 & 51.2 & 0.047 & 59.6 & \textbf{0.049} & \textbf{40.1} & \textbf{0.031} \\
    \hline 
    SeFa & 68.2 & 0.062 & 59.2 & 0.059 & 69.4 & 0.063 & \textbf{49.2} & 0.045 & 64.3 & 0.051 & 44.1 & 0.038 \\
    \hline 
    IGAN. & \textbf{48.9} & \textbf{0.034} & 59.6 & 0.059 & 62.3 & 0.056 & 53.1 & \textbf{0.039} & \textbf{58.8} & 0.053 & 45.8 & 0.039 \\
    \hline 
    SFlow. & 52.1 & 0.047 & 59.1 & 0.058 & 56.3 & \textbf{0.051} & 50.8 & 0.041 & 60.5 & 0.050 & 40.4 & 0.034 \\
    \hline 
    Ours & 51.2 & 0.048 & \textbf{58.8} & \textbf{0.057} & 60.8 & 0.053 & 50.3 & 0.041 & 61.1 & 0.051 & 42.1 & 0.035 \\
    \hline 
  \end{tabular}
}
\caption{Quantitative comparison of FID and KID among different image editing methods for \pigan{}, MVCGAN, and EG3D on attribute edits of face and animal images. The selected attributes are mentioned in ~\cref{sec:quan_details}. \label{tab:dist_level_metrics}} 

\end{table*}

\begin{figure}[!ht]
    \centering
    \includegraphics[width=.98\linewidth]{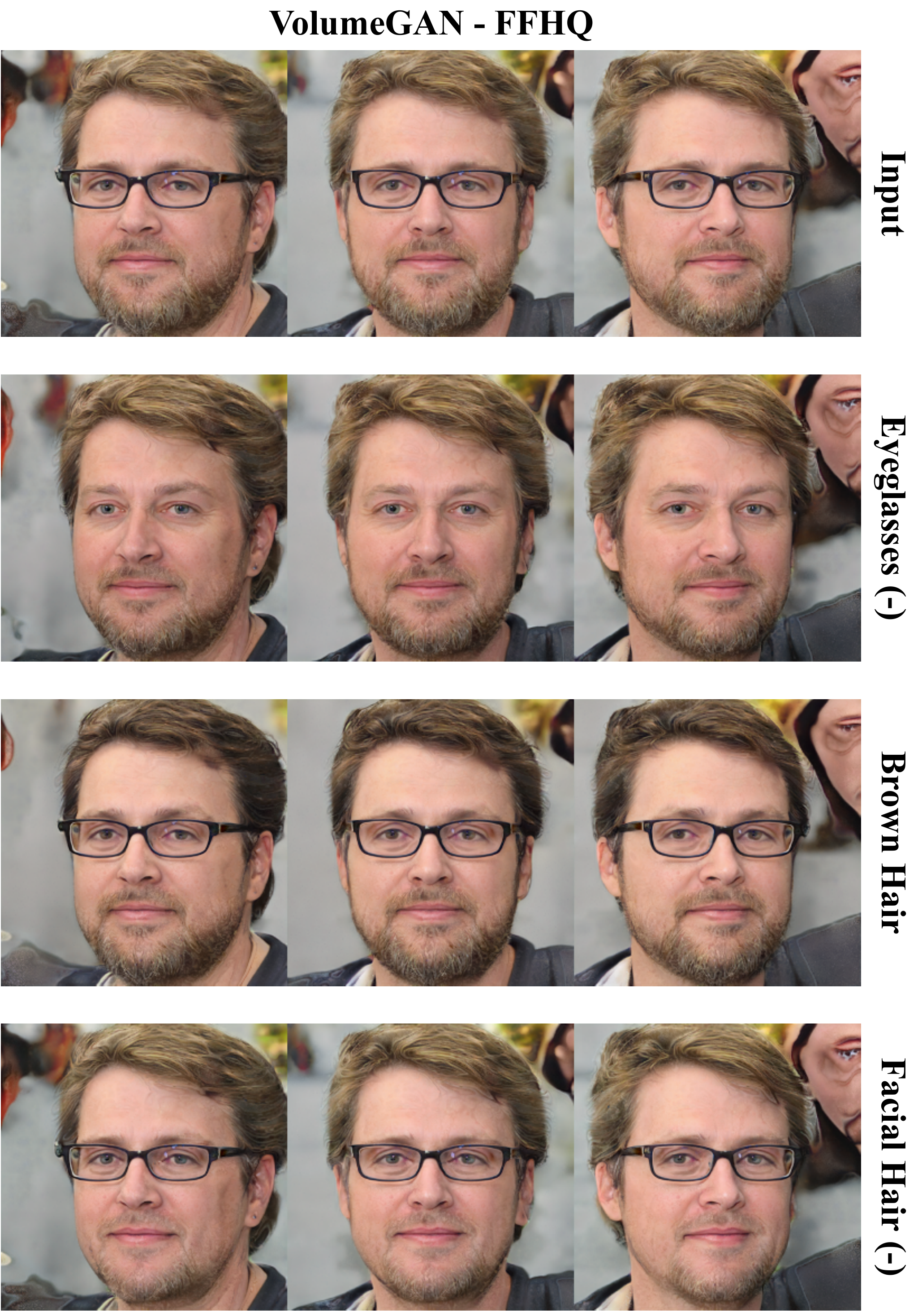}
    \caption{LatentSwap3D on VolumeGAN~\cite{xu2021volumegan} - FFHQ.}
    \label{fig:VolumeGAN_ffhq}
\end{figure}

\noindent\textbf{VolumeGAN} is a high-quality 3D-aware generative model explicitly trained to learn a structural and a textural representation, and it is based on NeRF~\cite{xu2021volumegan}. The results of our approach on VolumeGAN - FFHQ are provided in~\cref{fig:VolumeGAN_ffhq}. Our approach applies the desired attributes, \eg{,} removing eyeglasses, changing the hair color, and reducing the facial hair, to the latent space of VolumeGAN, without changing the identity of the input face.

\begin{figure}[!ht]
\vspace{0.2cm}
    \centering
    \includegraphics[width=.98\linewidth]{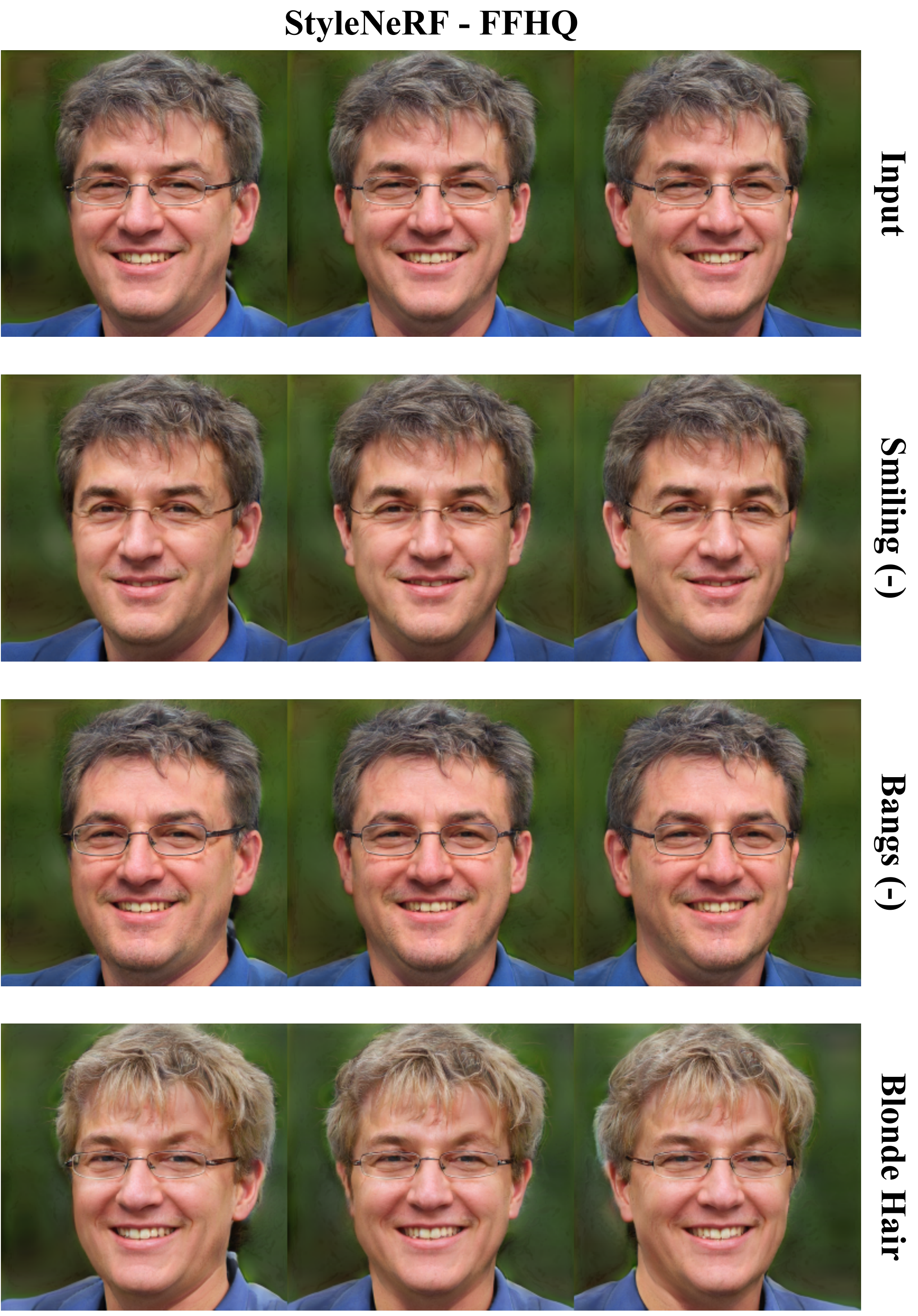}
    \caption{LatentSwap3D on StyleNeRF~\cite{gu2021stylenerf} - FFHQ.}
    \label{fig:stylenerf}
\end{figure}

\noindent\textbf{StyleNeRF} is another high-resolution 3D-aware generative model that integrates a NeRF into a 2D style-based generator~\cite{gu2021stylenerf}. StyleNeRF is able to generate high-resolution and 3D consistent images/shapes from unstructured 2D images. \Cref{fig:stylenerf} shows our attribute editing, \eg{,} smiling, removing bangs, and changing the hair color on StyleNeRF - FFHQ. \model{} operates successfully on the latent space of StyleNeRF by preserving the identity.

\subsection{Additional Quantitative Analysis}

We report \textit{Distribution-level Image Quality} metrics in addition to \textit{Identity Preservation} and \textit{Semantic Correctness} metrics.
We calculate Frechet Inception Distance (FID)~\cite{heusel2017gans,Seitzer2020FID} and Kernel Inception Distance (KID)~\cite{binkowski2018demystifying} between 10K edited images and the CelebA test dataset for the face domain and AFHQ test dataset for the animal domain. Although our method does not have the best FID and KID metrics, it is on par with other methods. Since these metrics do not indicate the methods' editing capability~\cite{ling2021editgan}, results in Tab.~\ref{tab:dist_level_metrics} should be considered together with \emph{semantic correctness} and \emph{identity preservation} metrics in Sec.~\ref{sec:quantitative}, Tab.~\ref{tab:semantic_correctness}, and Tab.~\ref{tab:identity_preservation} which instead measure the effectiveness of the edits. \model{} outperforms competitors in those metrics while providing competitive FID.

\section{Ablation Studies}\label{sec:additional_ablation}
This section reports some ablation studies on a study on hyper-parameters of \model{}, alternative latent space edit methods: linear operations and direct optimization of the desired attribute, a comparison of the use of random forests vs. other ranking methods for latent dimensions ranking, and camera pose optimization during inversion. %

\subsection{Hyper-Parameters}
\textbf{Selection of training set size.} We conduct an ablation study to assess the impact of training set size on a random forest that identifies the relevant dimensions for the given attribute. To this end, we generated additional samples from pre-trained generators at no extra cost. More samples will provide more diversity in the training set of the random forests; as expected, increasing the size of the training sets improves semantic correctness and identity preservation. However, doubling the training from 10K to 20K samples has diminishing returns (same \emph{Semantic Correctness}), as seen in Tab.~\ref{tab:rebuttal_table}. For the sake of efficiency, we picked 10K for all our experiments.

\begin{table}[!ht]

\centering
\begin{tabular}{|l|c|c|}
\hline
Training Set Size &  Sem. Cor. $\uparrow$ & Ident. Pres. $\uparrow$ \\
\hline

5K samples  & 92\%  & 73\%  \\ 
\hline
10K samples (default) & 95\%  & 71\%  \\ 
\hline
20K samples & 95\%  & 70\%  \\
\hline
\end{tabular}

\caption{\label{tab:rebuttal_table} Semantic correctness and Identity preservation metrics on various training set sizes on MVCGAN - FFHQ.} 

\end{table}

\textbf{Selection of $\tau$.} The metrics for \emph{Semantic Correctness} and \emph{Identity Preservation} with respect to different values of $\tau$ are presented in Tab.~\ref{tab:tau}. It is important to note that the choice of $\tau$ significantly impacts the identity preservation metric. Specifically, as $\tau$ increases, the identity preservation metric decreases. In contrast, the \emph{Semantic Correctness} metric exhibits diminishing returns after $\tau = 25\%$. Therefore, we picked this value and kept it constant during our experiments.

\begin{table}[!ht]

\centering
\begin{tabular}{|l|c|c|}
\hline
$\tau$ &  Sem. Cor. $\uparrow$ & Ident. Pres. $\uparrow$ \\
\hline

$15\%$ & 78\% & 88\% \\ 
\hline
$25\%$ & 95\% & 71\% \\ 
\hline
$35\%$ & 96\% & 65\% \\ 
\hline
$45\%$ & 97\% & 58\% \\ 
\hline
$55\%$ & 97\% & 44\% \\ 
\hline
$65\%$ & 97\% & 31\% \\ 
\hline
\end{tabular}  

\caption{Semantic correctness and Identity preservation metrics on different values of $\tau$ on MVCGAN - FFHQ.
\label{tab:tau} } 

\end{table}

\textbf{Selection of support set size.} We also report the semantic correctness and identity preservation metrics for different support set sizes for finding the reference image. As shown in Tab.~\ref{tab:support_set}, decreasing the support set size leads to improved semantic correctness since the reference image has more representative features based on the desired attribute. Conversely, increasing the support set size improves the identity loss metric. Based on the results, the optimal support set size identified is 32, which we used during all our experiments.

\begin{table}[!ht]

\centering
\begin{tabular}{|l|c|c|}
\hline
Support Set Size & Sem. Cor. $\uparrow$ & Ident. Pres. $\uparrow$ \\
\hline

$1$ & 95\% & 70\% \\ 
\hline
$16$ & 95\% & 70\% \\ 
\hline
$32$ & 95\% & 71\% \\ 
\hline
$128$ & 94\% & 72\% \\ 
\hline
$1024$ & 93\% & 72\% \\ 
\hline
\end{tabular}

\caption{Semantic correctness and Identity preservation metrics on different support set sizes on MVCGAN - FFHQ.
\label{tab:support_set} } 

\end{table}

\subsection{Alternative Edit Techniques}

\begin{figure}[ht!]

\centering
\scriptsize

\rotatebox[origin=lc]{90}{\centering \hspace{-0.5cm} \textbf{\phantom{y}}}
\begin{minipage}{.18\linewidth}
    \centering
    \phantom{\textbf{Input}}
\end{minipage}
\hfill\hfill
\begin{minipage}{.36\linewidth}
    \centering
    \textbf{Black Hair (+)}
\end{minipage}
\hfill\hfill
\begin{minipage}{.36\linewidth}
    \centering
    \textbf{Smiling (+)}
\end{minipage}

\rotatebox[origin=lc]{90}{\centering \hspace{-0.5cm} \textbf{\phantom{y}}}
\begin{minipage}{.18\linewidth}
    \centering
    \textbf{Input}
\end{minipage}
\hfill\hfill
\begin{minipage}{.18\linewidth}
    \centering
    \textbf{LM}
\end{minipage}
\begin{minipage}{.18\linewidth}
    \centering
    \textbf{Ours}
\end{minipage}
\hfill\hfill
\begin{minipage}{.18\linewidth}
    \centering
    \textbf{LM}
\end{minipage}
\begin{minipage}{.18\linewidth}
    \centering
    \textbf{Ours}
\end{minipage}

\vspace{0.1cm}

\rotatebox[origin=lc]{90}{\centering \hspace{-0.5cm} \textbf{\pigan{}\phantom{y}}}
\begin{minipage}{.18\linewidth}
    \centering
    \includegraphics[width=\linewidth]{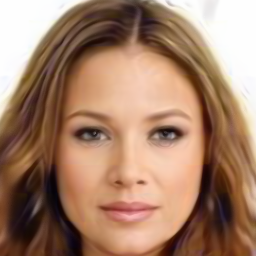}
\end{minipage}
\hfill\vline\hfill
\begin{minipage}{.18\linewidth}
    \centering
    \includegraphics[width=\linewidth]{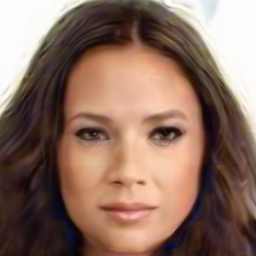}
\end{minipage}
\begin{minipage}{.18\linewidth}
    \centering
    \includegraphics[width=\linewidth]{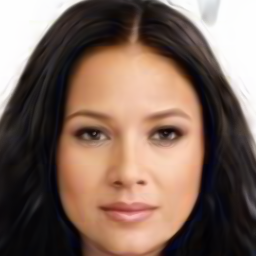}
\end{minipage}
\hfill\vline\hfill
\begin{minipage}{.18\linewidth}
    \centering
    \includegraphics[width=\linewidth]{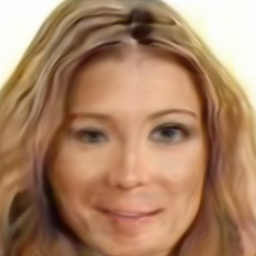}
\end{minipage}
\begin{minipage}{.18\linewidth}
    \centering
    \includegraphics[width=\linewidth]{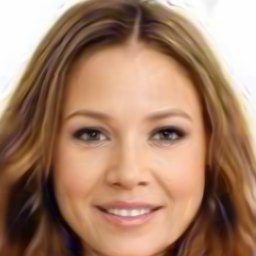}
\end{minipage}

\rotatebox[origin=lc]{90}{\centering \hspace{-0.55cm} \textbf{StyleSDF}}
\begin{minipage}{.18\linewidth}
    \centering
    \includegraphics[width=\linewidth]{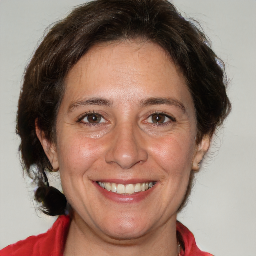}
\end{minipage}
\hfill\vline\hfill
\begin{minipage}{.18\linewidth}
    \centering
    \includegraphics[width=\linewidth]{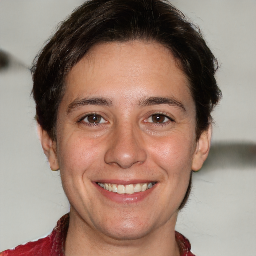}
\end{minipage}
\begin{minipage}{.18\linewidth}
    \centering
    \includegraphics[width=\linewidth]{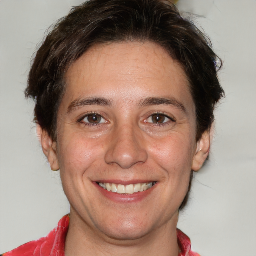}
\end{minipage}
\hfill\vline\hfill
\begin{minipage}{.18\linewidth}
    \centering
    \includegraphics[width=\linewidth]{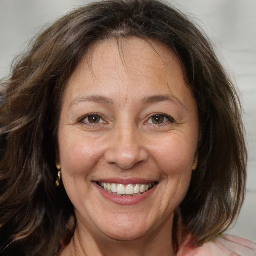}
\end{minipage}
\begin{minipage}{.18\linewidth}
    \centering
    \includegraphics[width=\linewidth]{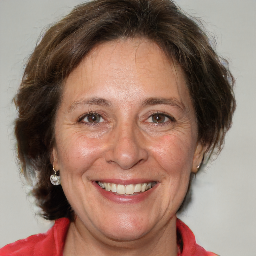}
\end{minipage}

\vspace{0.1cm}

\rotatebox[origin=lc]{90}{\centering \hspace{-0.5cm} \textbf{\phantom{y}}}
\begin{minipage}{.18\linewidth}
    \centering
    \textbf{Input}
\end{minipage}
\hfill\hfill
\begin{minipage}{.18\linewidth}
    \centering
    \textbf{LM}
\end{minipage}
\begin{minipage}{.18\linewidth}
    \centering
    \textbf{Ours}
\end{minipage}
\hfill\hfill
\begin{minipage}{.18\linewidth}
    \centering
    \textbf{LM}
\end{minipage}
\begin{minipage}{.18\linewidth}
    \centering
    \textbf{Ours}
\end{minipage}

\rotatebox[origin=lc]{90}{\centering \hspace{-0.5cm} \textbf{\phantom{y}}}
\begin{minipage}{.18\linewidth}
    \centering
    \phantom{\textbf{Input}}
\end{minipage}
\hfill\hfill
\begin{minipage}{.36\linewidth}
    \centering
    \textbf{Female (-)}
\end{minipage}
\hfill\hfill
\begin{minipage}{.36\linewidth}
    \centering
    \textbf{Age (+)}
\end{minipage}

\caption{Effect of \emph{linear manipulations} (LM) on the latent space of the 3D GANs, \pigan{} and StyleSDF.\label{fig:linear_operations}}
\end{figure}

\paragraph{Linear operations.} StyleGAN-based generators use AdaIN~\cite{huang2017arbitrary} layers to guide the image generation process. AdaIN layers apply a linear transformation to the input features; therefore, they are suitable to be modified with simple linear transformations in the latent space. For example, recently~\cite{wu2021stylespace} showed that such linear operations are enough to provide disentangled and fine-grained manipulations in the latent space of StyleGAN2~\cite{Karras2020stylegan2}. 
While some 3D GANs employ a style space where linear operations can be applied, such as EG3D, others do not, and one of the objectives of this work was to be able to develop a method that is completely generator agnostic.
For example, \pigan{} and StyleSDF use SIREN~\cite{sitzmann2020siren} layers that enforce periodicity due to the presence of $\sin$-based activation functions in the learned latent space.
Intuitively linear edits of latent codes (such as additions or subtractions) will not perform nicely in a periodic latent space, therefore motivating the need to resort to the feature swapping mechanism of \model{}.
To verify this intuition, we conducted an ad-hoc experiment performing linear operations, such as addition and subtraction, on the latent spaces of \pigan{} and StyleSDF. We reported the results in \cref{fig:linear_operations}. Linear edits in this context are defined as constant changes on the top 256 features ranked from our trained Random Forests. To increase or decrease the corresponding latent codes, we look at the sign of the difference between the latent codes of the image that will be edited and the reference images that have the desired attribute. 
Linear operations on \pigan{} can result in images with some artefacts (\eg{,} \pigan{} - smiling) or lower intensity edits (\eg{,} \pigan{} - black hair). 
For StyleSDF, linear operations can apply the desired edit and generate realistic images but have an undesired side effect on the identity.
The StyleSDF - female edit changes the background and clothing, while the StyleSDF - age edit also changes the hairstyle. 
For both generators and all four attributes, \model{} can perform disentangled edits free of artifacts and preserving identity.

\begin{figure}[ht!]
\centering

\begin{minipage}{.16\linewidth}
    \centering
    \scriptsize{\textbf{Original Image}}
\end{minipage}
\begin{minipage}{.075\linewidth}
    \raggedleft
    \phantom{\rotatebox[origin=lc]{90}{\centering \hspace{-0.3cm} \scriptsize{\textbf{-}}}}
\end{minipage}
\begin{minipage}{.16\linewidth}
    \centering
    \scriptsize{\textbf{100 iterations}}
\end{minipage}
\begin{minipage}{.16\linewidth}
    \centering
    \scriptsize{\textbf{200 iterations}}
\end{minipage}
\begin{minipage}{.16\linewidth}
    \centering
    \scriptsize{\textbf{300 iterations}}
\end{minipage}
\begin{minipage}{.16\linewidth}
    \centering
    \scriptsize{\textbf{400 iterations}}
\end{minipage}

\begin{minipage}{.16\linewidth}
    \centering
    \includegraphics[width=\linewidth]{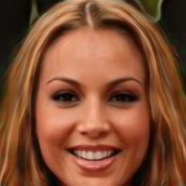}
\end{minipage}
\begin{minipage}{.08\linewidth}
    \raggedleft
    \rotatebox[origin=lc]{90}{\centering \scriptsize{\textbf{Smiling (+)}}}
\end{minipage}
\begin{minipage}{.64\linewidth}
    \centering
    \includegraphics[width=\linewidth]{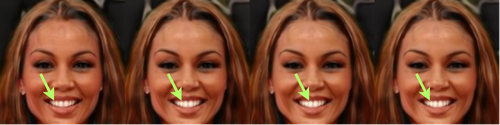}
\end{minipage}

\begin{minipage}{.16\linewidth}
    \centering
    \phantom{\includegraphics[width=\linewidth]{figures/attribute_optimization/original.png}}
\end{minipage}
\begin{minipage}{.08\linewidth}
    \raggedleft
    \rotatebox[origin=lc]{90}{\centering \scriptsize{\textbf{Smiling (-)}}}
\end{minipage}
\begin{minipage}{.64\linewidth}
    \centering
    \includegraphics[width=\linewidth]{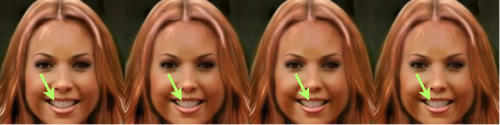}
\end{minipage}

\begin{minipage}{.16\linewidth}
    \centering
    \phantom{\includegraphics[width=\linewidth]{figures/attribute_optimization/original.png}}
\end{minipage}
\begin{minipage}{.08\linewidth}
    \raggedleft
    \rotatebox[origin=lc]{90}{\centering  \scriptsize{\textbf{\phantom{y}Black Hair}}}
\end{minipage}
\begin{minipage}{.64\linewidth}
    \centering
    \includegraphics[width=\linewidth]{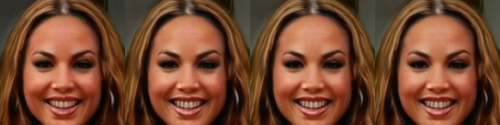}
\end{minipage}

\begin{minipage}{.16\linewidth}
    \centering
    \phantom{\includegraphics[width=\linewidth]{figures/attribute_optimization/original.png}}
\end{minipage}
\begin{minipage}{.08\linewidth}
    \raggedleft
    \rotatebox[origin=lc]{90}{\centering  \scriptsize{\textbf{\phantom{y}Blond Hair}}}
\end{minipage}
\begin{minipage}{.64\linewidth}
    \centering
    \includegraphics[width=\linewidth]{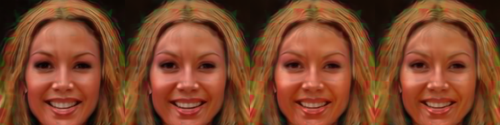}
\end{minipage}

\caption{Directly optimize the latent codes using pre-trained image classifiers and back-propagation. \label{fig:attribute_optimization}}
\vspace{-0.4cm}
\end{figure}

\paragraph{Direct optimization.} Since we have differentiable image classifiers for the attribute we would like to edit, an alternative to \model{} would be to directly optimize latent codes to maximize the presence of the desired attribute as measured by the respective classifiers. 
To compare against this alternative, we first trained binary image classifiers based on ResNet-50~\cite{resnet50} on the CelebA~\cite{liu2015faceattributes} dataset for each attribute. 
Then, we take the latent codes of the original image as the initial point and try to learn an offset in the latent space that, when summed to the initial latent code, applies the desired transformation. To optimize the offset, we feed the generator the original latent code modified by the offset, generate an edited image, and provide it to the attribute classifier. At this point, we can compute as a loss function the cross-entropy loss between the output of the classifier and the class of the desired attribute and minimize it to optimize the offset using back-propagation directly. For the optimization procedure, we perform 400 iterations with Adam~\cite{kingma2014adam}. As shown in \cref{fig:attribute_optimization}, this method struggles to preserve the identity of the edited image (see all lines). Moreover, it learns edits that are not realistic but are classifier-biased, such as the smiling (+) attribute that brightens the teeth. In contrast, the smiling (-) attribute changes the color of the teeth to the skin color; see light-green arrows in \cref{fig:attribute_optimization}.

\subsection{Other Feature Ranking Methods}
\begin{figure}[ht!]
\centering
\scriptsize

\begin{minipage}{.16\linewidth}
    \centering
    \scriptsize{\textbf{Input}}
\end{minipage}
\begin{minipage}{.075\linewidth}
    \raggedleft
    \phantom{\rotatebox[origin=lc]{90}{\centering \hspace{-0.3cm} \scriptsize{\textbf{-}}}}
\end{minipage}
\begin{minipage}{.16\linewidth}
    \centering
    \scriptsize{\textbf{Gender}}
\end{minipage}
\begin{minipage}{.16\linewidth}
    \centering
    \scriptsize{\textbf{Smiling (+)}}
\end{minipage}
\begin{minipage}{.16\linewidth}
    \centering
    \scriptsize{\textbf{Blonde (+)}}
\end{minipage}
\begin{minipage}{.16\linewidth}
    \centering
    \scriptsize{\textbf{Makeup (+)}}
\end{minipage}

\begin{minipage}{.16\linewidth}
    \centering
    \includegraphics[width=\linewidth]{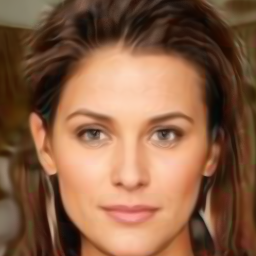}
\end{minipage}
\begin{minipage}{.08\linewidth}
    \raggedleft
    \rotatebox[origin=lc]{90}{\scriptsize{\textbf{SVM}}}
\end{minipage}
\begin{minipage}{.16\linewidth}
    \centering
    \includegraphics[width=\linewidth]{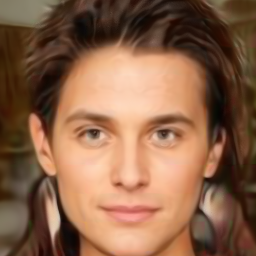}
\end{minipage}
\begin{minipage}{.16\linewidth}
    \centering
    \includegraphics[width=\linewidth]{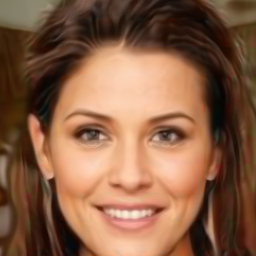}
\end{minipage}
\begin{minipage}{.16\linewidth}
    \centering
    \includegraphics[width=\linewidth]{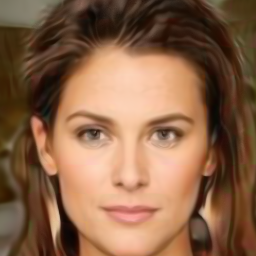}
\end{minipage}
\begin{minipage}{.16\linewidth}
    \centering
    \includegraphics[width=\linewidth]{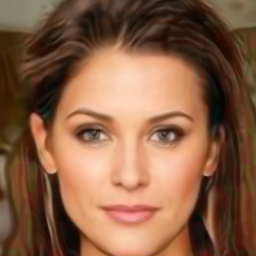}
\end{minipage}

\begin{minipage}{.16\linewidth}
    \centering
    \phantom{\includegraphics[width=\linewidth]{figures/different_ranking_methods/2103843295_original.png}}
\end{minipage}
\begin{minipage}{.08\linewidth}
    \raggedleft
    \rotatebox[origin=lc]{90}{\scriptsize{\textbf{SelectKBest}}}
\end{minipage}
\begin{minipage}{.16\linewidth}
    \centering
    \includegraphics[width=\linewidth]{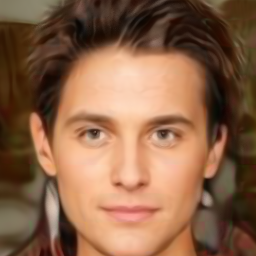}
\end{minipage}
\begin{minipage}{.16\linewidth}
    \centering
    \includegraphics[width=\linewidth]{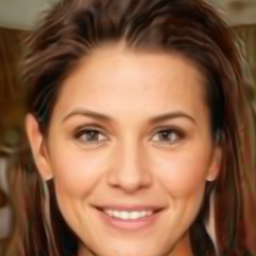}
\end{minipage}
\begin{minipage}{.16\linewidth}
    \centering
    \includegraphics[width=\linewidth]{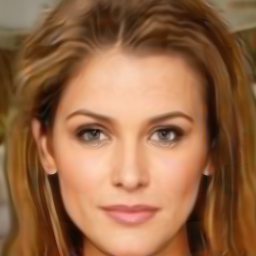}
\end{minipage}
\begin{minipage}{.16\linewidth}
    \centering
    \includegraphics[width=\linewidth]{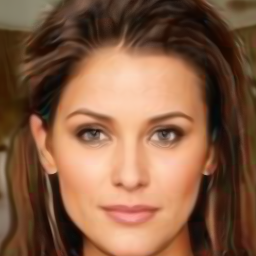}
\end{minipage}

\begin{minipage}{.16\linewidth}
    \centering
    \phantom{\includegraphics[width=\linewidth]{figures/different_ranking_methods/2103843295_original.png}}
\end{minipage}
\begin{minipage}{.08\linewidth}
    \raggedleft
    \rotatebox[origin=lc]{90}{\scriptsize{\textbf{SHAP}}}
\end{minipage}
\begin{minipage}{.16\linewidth}
    \centering
    \includegraphics[width=\linewidth]{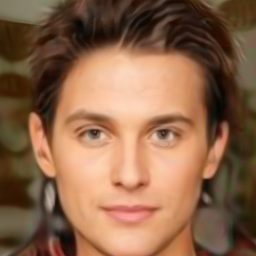}
\end{minipage}
\begin{minipage}{.16\linewidth}
    \centering
    \includegraphics[width=\linewidth]{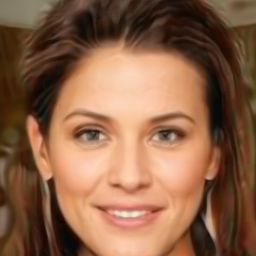}
\end{minipage}
\begin{minipage}{.16\linewidth}
    \centering
    \includegraphics[width=\linewidth]{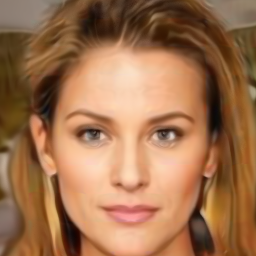}
\end{minipage}
\begin{minipage}{.16\linewidth}
    \centering
    \includegraphics[width=\linewidth]{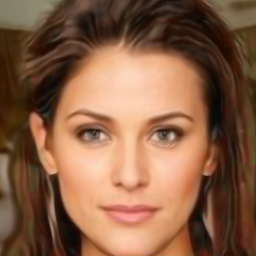}
\end{minipage}

\begin{minipage}{.16\linewidth}
    \centering
    \phantom{\includegraphics[width=\linewidth]{figures/different_ranking_methods/2103843295_original.png}}
\end{minipage}
\begin{minipage}{.08\linewidth}
    \raggedleft
    \rotatebox[origin=lc]{90}{ \scriptsize{\textbf{RF (used)}}}
\end{minipage}
\begin{minipage}{.16\linewidth}
    \centering
    \includegraphics[width=\linewidth]{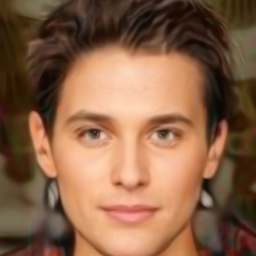}
\end{minipage}
\begin{minipage}{.16\linewidth}
    \centering
    \includegraphics[width=\linewidth]{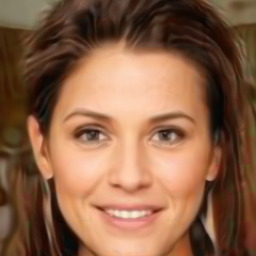}
\end{minipage}
\begin{minipage}{.16\linewidth}
    \centering
    \includegraphics[width=\linewidth]{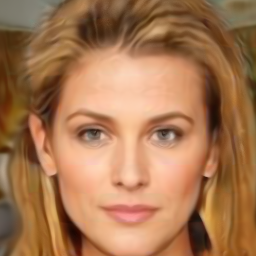}
\end{minipage}
\begin{minipage}{.16\linewidth}
    \centering
    \includegraphics[width=\linewidth]{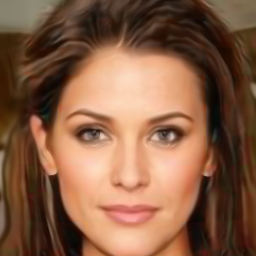}
\end{minipage}

\caption{Comparison of various feature ranking methods on the latent space of \pigan{}.\label{fig:ranking_methods}}
\vspace{-0.4cm}
\end{figure}

In \model{}, we use Random Forests~\cite{breiman2001random} based feature ranking; in this section, we experimentally motivate this choice by considering other methods for feature ranking.
We consider three alternatives: (i) the \textit{SelectKBest}~\cite{scikit-learn} method from the popular SciKit-Learn library that sorts the feature based on a score function, such as $\chi^2$~\cite{pearson1900x}, and selects the $k$ features with the highest scores, (ii) \textit{Support Vector Machine} (SVM)~\cite{cortes1995support} based method that takes the absolute values of the feature coefficients of a trained linear SVM, and (iii) \textit{SHapley Additive exPlanations} (SHAP)~\cite{lundberg2020local2global} based method that explains the output of trained machine learning models by calculating the importance of the features. As can be seen in \cref{fig:ranking_methods}, SHAP- and RF-based methods show similar performance on the attributes \textit{female (-), smiling (+)}. However, for \textit{blondness (+) and makeup (+)} attribute edits, the random forest-based ranking provides higher quality. On the other hand, SVM-based ranking has comparable results for \textit{smiling (+) and makeup (+)}, but for the other attributes fails to generate the corresponding edits. Finally, the \textit{SelectKBest} method performs similarly to the SVM-based ranking method, but it has a small effect on the blondness attribute.

\subsection{Off-the-shelf Inversion Method}

\begin{figure}[ht!]
\centering
\scriptsize
\vspace{-0.25cm}

\begin{minipage}{.235\linewidth}
    \centering
    \textbf{Input}
\end{minipage}
\begin{minipage}{.235\linewidth}
    \centering
    \textbf{Inversion (14\%)}
\end{minipage}
\begin{minipage}{.235\linewidth}
    \centering
    \textbf{Smiling (+) (20\%)}
\end{minipage}

\begin{minipage}{.235\linewidth}
    \centering
    \includegraphics[width=\linewidth]{figures/teaser/latentswap3d/gt.jpg}
\end{minipage}
\begin{minipage}{.235\linewidth}
    \centering
    \includegraphics[width=\linewidth]{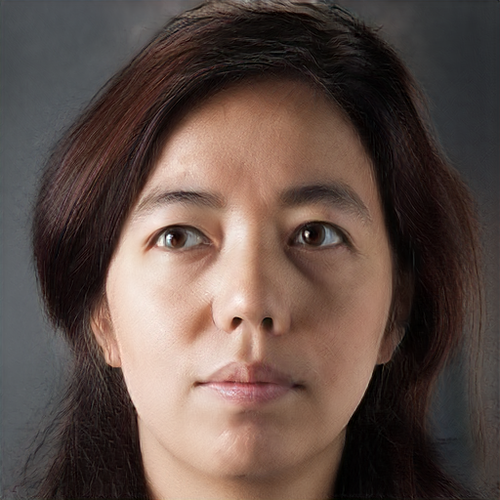}
\end{minipage}
\begin{minipage}{.235\linewidth}
    \centering
    \includegraphics[width=\linewidth]{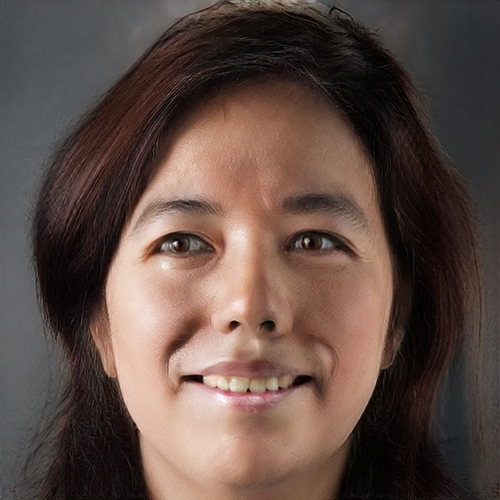}
\end{minipage}

\caption{Percentage (\%) denotes the identity change from the input image.}
\label{fig:sota_inv}
\end{figure}

Though the central objective of this paper does not revolve around proposing an inversion technique, we illustrate a use-case scenario using our proposed method for real image editing in the main paper.
Figure~\ref{fig:sota_inv} shows the combination of our work with a SoTA inversion method\cite{yin20233d} for EG3D.  
Our method remains valid and applicable, combined with arbitrary GAN inversion methods (including the most recent ones).

\subsection{Consecutive and Complex Edits}
\begin{figure}[ht!]
\centering
\scriptsize

\begin{minipage}{.235\linewidth}
    \centering
    \textbf{Input}
\end{minipage}
\begin{minipage}{.235\linewidth}
    \centering
    \textbf{(1) Blue Eyes (7\%)}
\end{minipage}
\begin{minipage}{.235\linewidth}
    \centering
    \textbf{(2) Smiling (15\%)}
\end{minipage}

\begin{minipage}{.235\linewidth}
    \centering
    \includegraphics[width=\linewidth]{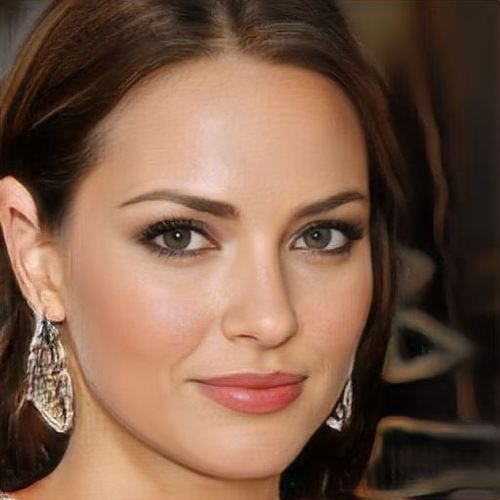}
\end{minipage}
\begin{minipage}{.235\linewidth}
    \centering
    \includegraphics[width=\linewidth]{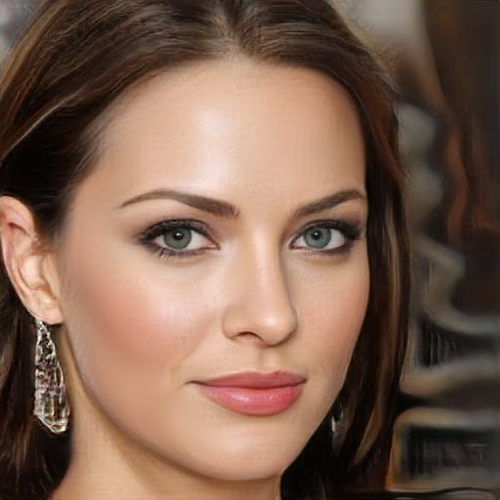}
\end{minipage}
\begin{minipage}{.235\linewidth}
    \centering
    \includegraphics[width=\linewidth]{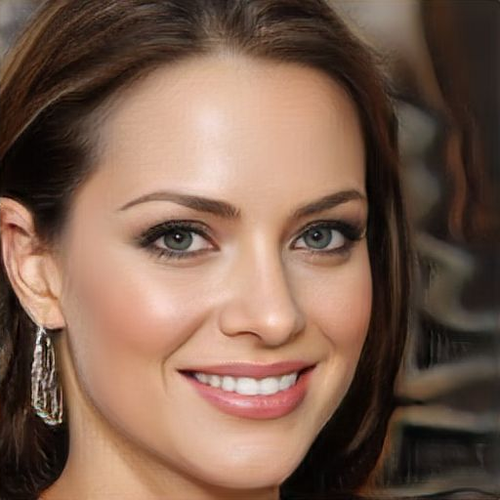}
\end{minipage}

\begin{minipage}{.235\linewidth}
    \centering
    \includegraphics[width=\linewidth]{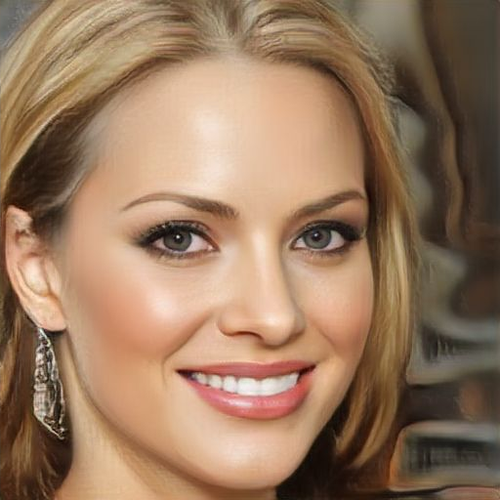}
\end{minipage}
\begin{minipage}{.235\linewidth}
    \centering
    \includegraphics[width=\linewidth]{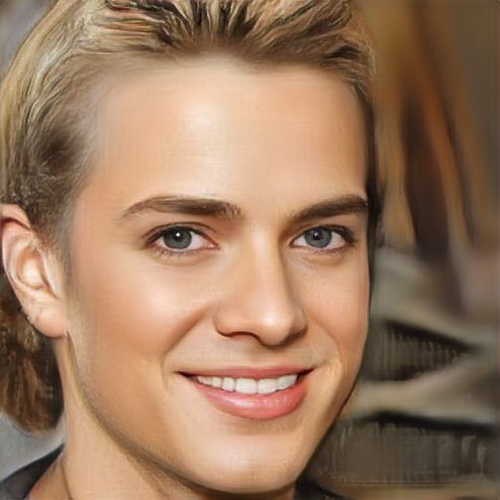}
\end{minipage}
\begin{minipage}{.235\linewidth}
    \centering
    \includegraphics[width=\linewidth]{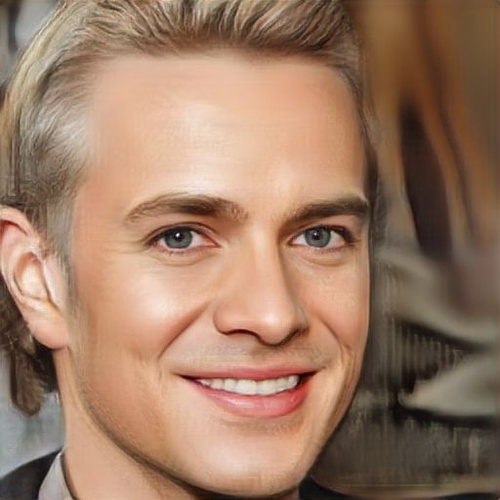}
\end{minipage}

\begin{minipage}{.235\linewidth}
    \centering
    \textbf{(3) Blonde (17\%)}
\end{minipage}
\begin{minipage}{.235\linewidth}
    \centering
    \textbf{(4) Gender (28\%)}
\end{minipage}
\begin{minipage}{.235\linewidth}
    \centering
    \textbf{(5) Age (+) (22\%)}
\end{minipage}

\caption{We  report five sequential edits, where \% denotes the identity change between consecutive edits.}
\label{fig:cumulative_edits}
\end{figure}

We provide consecutive edits in Fig.~\ref{fig:cumulative_edits}. \model{} operates multiple edits, such as \textit{blue eyes}, \textit{smiling}, \textit{blonde}, \textit{gender}, and \textit{age}.

\subsection{Details on Camera Pose Optimization}
\begin{figure}[!ht]
    \centering
    \scriptsize
    
    \begin{minipage}{0.16\linewidth}
        \centering
        \textbf{Input}
    \end{minipage}
    \begin{minipage}{0.80\linewidth}
        \centering
        \textbf{Different Views}
    \end{minipage}
    \rotatebox[origin=lc]{270}{\centering \phantom{\textbf{a}}}
    
    \begin{minipage}{0.16\linewidth}
        \centering
        \includegraphics[width=\linewidth]{figures/teaser/latentswap3d/gt.jpg}
    \end{minipage}
    \begin{minipage}{0.80\linewidth}
        \centering
        \includegraphics[width=\linewidth]{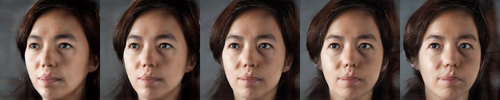}
    \end{minipage}
    \rotatebox[origin=lc]{270}{\centering \hspace{-0.6cm} \textbf{w/out opt.}}
    
    \begin{minipage}{0.16\linewidth}
        \centering
        \phantom{\includegraphics[width=\linewidth]{figures/teaser/latentswap3d/gt.jpg}}
    \end{minipage}
    \begin{minipage}{0.80\linewidth}
        \centering
        \includegraphics[width=\linewidth]{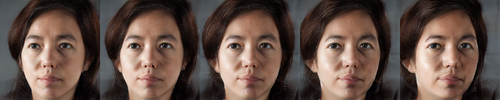}
    \end{minipage}
    \rotatebox[origin=lc]{270}{\centering \hspace{-0.35cm} \textbf{Ours\phantom{p}}}
    
    \begin{minipage}{0.16\linewidth}
        \centering
        \phantom{\includegraphics[width=\linewidth]{figures/teaser/latentswap3d/gt.jpg}}
    \end{minipage}
    \begin{minipage}{0.80\linewidth}
        \centering
        \includegraphics[width=\linewidth]{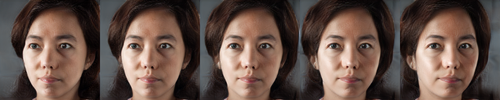}
    \end{minipage}
    \rotatebox[origin=lc]{270}{\centering \hspace{-0.25cm} \textbf{\cite{deng2019accurate}\phantom{p}}}

    \caption{Importance of Camera Optimization in Inversion Procedure. Off-the-shelf pose estimator~\cite{deng2019accurate}.}
    \label{fig:cam_optimization}
    \vspace{-0.5cm}

\end{figure}

Using off-the-shelf face pose estimation can be an alternative to the proposed method for the specific case of faces. However, it will hinder the generalizability of the inversion procedure to only those datasets or object categories for which a pose estimator can be trained. Our alternating optimization schema, instead, only relies on the assumption of having a trained generator and, as such, we believe, provides a more general solution. To show the impact of optimizing the pose on the inversion process, we report a comparison in~\cref{fig:cam_optimization}.

\section{Limitations}\label{section:limitations}
\begin{figure}[ht!]
\centering
\scriptsize

\rotatebox[origin=lc]{90}{\centering \hspace{-0.5cm} \textbf{\phantom{y}}}
\begin{minipage}{.18\linewidth}
    \centering
    \textbf{Input}
\end{minipage}
\begin{minipage}{.54\linewidth}
    \centering
    \textbf{Edited Image}
\end{minipage}
\begin{minipage}{.18\linewidth}
    \centering
    \textbf{Reference}
\end{minipage}

\vspace{0.05cm}

\rotatebox[origin=lc]{90}{\centering \hspace{-0.65cm} \textbf{Earrings}}
\begin{minipage}{.18\linewidth}
    \centering
    \includegraphics[width=\linewidth]{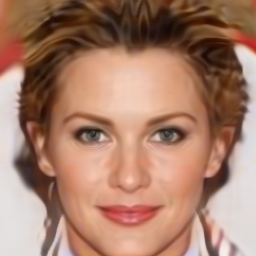}
\end{minipage}
\begin{minipage}{.54\linewidth}
    \centering
    \includegraphics[width=\linewidth]{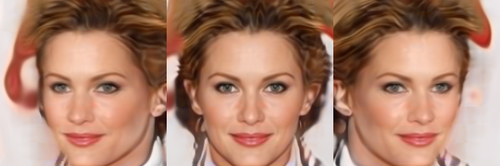}
\end{minipage}
\begin{minipage}{.18\linewidth}
    \centering
    \includegraphics[width=\linewidth]{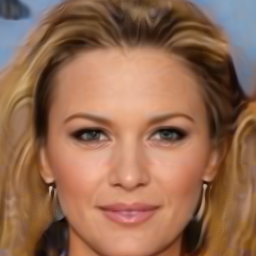}
\end{minipage}

\rotatebox[origin=lc]{90}{\centering \hspace{-0.4cm} \textbf{Hat\phantom{y}}}
\begin{minipage}{.18\linewidth}
    \centering
    \includegraphics[width=\linewidth]{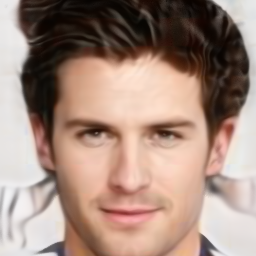}
\end{minipage}
\begin{minipage}{.54\linewidth}
    \centering
    \includegraphics[width=\linewidth]{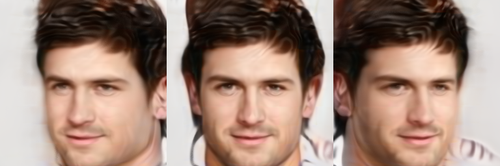}
\end{minipage}
\begin{minipage}{.18\linewidth}
    \centering
    \includegraphics[width=\linewidth]{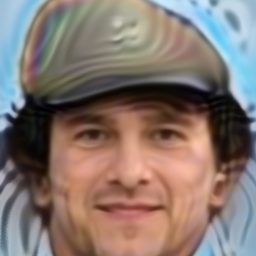}
\end{minipage}

\caption[Failure Cases]{Failure edits on the latent space of \pigan{} for classes under-represented in the training set. \label{fig:failure_cases}}
\vspace{-0.55cm}

\end{figure}

\paragraph{Under-represented Attributes on Training Datasets of GANs.} During the development of this work, we identified some attribute manipulations that cannot be applied in the latent space of pre-trained 3D-aware image generators. 
These usually cover under-represented classes in the original training set, such as faces with a hat or earrings. 
We hypothesize that these samples fall out of distribution for the generator, so they do not have specific dimensions in the latent space allocated to them. For this reason, reproducing them with our editing technique is difficult. We show some failed edits in \cref{fig:failure_cases}.  

\begin{figure}[ht!]
\centering
\scriptsize
\begin{minipage}{.16\linewidth}
    \centering
    \textbf{Input}
\end{minipage}
\begin{minipage}{.80\linewidth}
    \centering
    \textbf{Different Views}
\end{minipage}

\begin{minipage}{.16\linewidth}
    \centering
    \includegraphics[width=\linewidth]{figures/pigan/inversion/merkel.png}
\end{minipage}
\begin{minipage}{.80\linewidth}
    \centering
    \includegraphics[width=\linewidth]{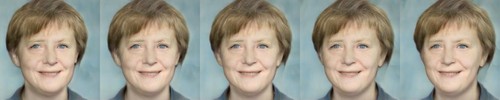}
\end{minipage}

\caption{Inversion on the latent space of StyleSDF. While the global appearance matches, the identity is not preserved.
\label{fig:stylesdf_inversion}}
\vspace{-0.9cm}
\end{figure}

\paragraph{Real Images Inverting Capabilities of GANs.} We showed promising initial results on editing real images via GAN inversion followed by \model{}. 
During the development of this work, we found that the inversion of an image in the latent space of 3D generators is quite challenging and sometimes fails to generate high-quality outputs or maintain the identity of the inverted face. 
This is particularly true for StyleSDF, where the inverted faces resemble the original but not perfectly. We show one example inversion in \cref{fig:stylesdf_inversion}. 
However, this limitation is naturally solved using newer and more powerful generators with  better inversion capabilities, \eg{,} MVCGAN. As shown in the main manuscript, our model can produce consistent attribute editing on real images with a powerful generator.

\section{Implementation Details}\label{section:implementation_details}
\subsection{Datasets}\label{section:datasets}
We test our proposed model, \model{}, with 3D-aware generative models on images from six different datasets: CelebA (256x256)~\cite{liu2015faceattributes}, FFHQ (1024x1024)~\cite{karras2019style}, Cats (256x256)~\cite{cats}, AFHQ (512x512)~\cite{choi2020starganv2}, and CompCars~\cite{yang2015large} (256x256) method to the 2D GAN model StyleGAN2 on FFHQ (1024x1024)~\cite{karras2019style}, AFHQ (512x512)~\cite{choi2020starganv2}, and MetFaces (1024x1024)~\cite{Karras2020ada}. 

\noindent \textbf{CelebA}~\cite{liu2015faceattributes} is a large-scale dataset of $\sim$200K face images of over 10K different celebrities and 40 annotated attributes for each image. Its resolution is (256x256). \pigan{} and MVCGAN provide pre-trained weights for CelebA. 

\noindent \textbf{FFHQ} dataset consists of over 70K high-resolution~(1024x1024) and high-fidelity images of human faces~\cite{karras2019style}. The dataset has diverse samples regarding age, ethnicity, and wearing accessories. It is used in this work to generate images from StyleNeRF, VolumeGAN, GIRAFFE, StyleSDF, MVCGAN, and EG3D.

\noindent \textbf{Cats} contains 6K images~(128x128) of cat heads~\cite{cats}. The dataset is used to generate cat images from \pigan{}.

\noindent \textbf{AFHQ} is a dataset that contains over 15K high-resolution~(512x512) and high-quality images of animal faces~\cite{choi2020starganv2}. The dataset has three domains: dogs, cats, and wildlife animals, and each domain has 5K samples. %

\noindent \textbf{MetFaces} consists of over 1K human faces (1024x1024) extracted from works of art. The images are automatically aligned and cropped~\cite{Karras2020ada}. This dataset is used only when evaluating \model{} on StyleGAN2.

\noindent \textbf{CompCars} contains $\sim$137K images of 1716 unique car models~\cite{yang2015large}. Its resolution is 256x256. This dataset is used only when evaluating \model{} on GIRAFFE.

\subsection{Runtime Analysis}
\label{section:runtime_analysis}
\model{} consists of two main steps, as shown in Fig.~\ref{fig:step1} and~\ref{fig:step2}. We measure the runtime on an NVIDIA Tesla T4 GPU Machine with 12-cores. 

\begin{table}[!ht]
  \centering

  \begin{tabular}{|l|c|c|c|}
    \hline
     & Identifying & Editing & Image \\
    Method & Step & Step & Inversion \\
    \hline
    \pigan{} & 180 min. & 600 ms/im. & 20 min./im. \\ 
    \hline
    MVCGAN & 68 min. & 600 ms/im. & 8 min./im. \\ 
    \hline
    EG3D & 72 min. & 600 ms/im. & 8 min./im. \\ 
    \hline
  \end{tabular}

\caption{Overall runtime analysis of the proposed method. The values for the first column are calculated using a dataset of 10K generated images.\label{tab:runtime_analysis}} 
\end{table}

\paragraph{Identifying Relevant Latent Dimensions.} For the step in Sec.~\ref{sec:ranking}, there are three main processes: \textbf{(i)} generating the training set from random sampling in the latent space of the generator, \textbf{(ii)} predicting the probabilities of the presence of the desired attribute in the generated images using pre-trained image classifiers, and \textbf{(iii)} training a random forest to predict the presence of the desired attribute from the latent codes. Considering \pigan{} as a generator, the image generation step takes 2 hours for 10K images, while for MVCGAN and EG3D, it takes 8 and 12 minutes, respectively. Labeling the 10K images using the pre-trained image classifiers takes ~45 seconds per attribute. Finally, the training process of the random forests takes 1 minute per attribute.

\paragraph{Attribute Editing on Latent Dimensions.} The second step is described in Sec.~\ref{sec:editing}, which takes around 600 milliseconds per image for all generators. 

\paragraph{3D Edits on Real Images.} The runtime analysis for each generator's inversion of real images is shown in \cref{tab:runtime_analysis}. The inversion procedure can be sped up using encoder-based inversion approaches. However, we leave it to future development.

\subsection{Details of Animal Attribute Classifiers} \label{sec:animal_classifier}
We trained ResNet-50~\cite{resnet50} classifiers to predict \textit{Siamese breed} and \textit{brown color} by using the dataset~\cite{petfinder}. Since we do not have frontal and zoom-in views of the animals, we apply a haar detector\footnote{\url{https://github.com/kipr/opencv/blob/master/data/haarcascades/haarcascade\_frontalface\_default.xml}} for cat faces for the dataset. Our model can successfully edit AFHQ and Cats datasets by leveraging these attribute classifiers. 

\subsection{Details on the Quantitative Analysis} \label{sec:quan_details}
For \textit{Distribution-level Image Quality} and \textit{Identity preservation} metrics, we use 2000 generated images per attribute from five different attributes, 10K in total per method. For \model{}, InterFaceGAN~\cite{shen2020interfacegan}, and StyleFlow~\cite{abdal2021styleflow}, we select the attributes for the three generators as follows: for \pigan{} we tested \textit{gender, smile, age, hair color, and heavy makeup}, while for MVCGAN and EG3D, we picked \textit{gender, smile, age, glasses, and adding beard}. 
SeFa~\cite{shen2021closed} and LatentCLR~\cite{yuksel2021latentclr} are unsupervised edits discovery methods. Therefore we cannot isolate specific attribute editing transformations. So instead, we take the top five \textit{semantics} for SeFa and five \textit{directional models} for LatentCLR.

\subsection{Details on the Comparison to Other Methods.} 
Since the other 3D editing methods apply to specific architectures or have their own generator part, we pick 2D attribute manipulators that have been proven to work well on 2D generators as baselines. They can also be applied to latent spaces of 3D GANs.
InterFaceGAN~\cite{shen2020interfacegan} and StyleFlow~\cite{abdal2021styleflow} are the closest competitors to our method and were originally proposed for image generators. Similarly to \model{}, InterFaceGAN leverages pre-trained attribute classifiers to find the corresponding linear edit directions in the latent space of trained generators. However, as mentioned, linear edits are sub-optimal in the periodic space determined by the SIREN~\cite{sitzmann2020siren} activation functions used in \pigan{}, MVCGAN, and others. 
On the other hand, StyleFlow uses the attribute information during the training of normalizing flows as conditions. When editing the desired attribute on a face sample, the user can give the desired attribute as a condition. 
In our comparison, we also consider methods for unsupervised discovery of editing directions: SeFa~\cite{shen2021closed} and LatentCLR~\cite{yuksel2021latentclr}. Both methods do not have assumptions about the characteristic of the generator to which they are applied. Therefore, they can be easily adapted to NeRF-based generators like \pigan{} or MVCGAN.

\section{Future Work}\label{section:future_work}
\paragraph{Real Images Inverting Capabilities of GANs.} While a better 3D-aware GAN inversion was outside the scope of this work, we believe that in the future, some of the proposed techniques for style-based 2D generators like~\cite{roich2021pivotal} could be adapted for the new category of 3D-aware generators and combined with \model{} to enable even more powerful edits on real images. For instance, if encoder-based inversion~\cite{roich2021pivotal} is adapted, it will speed up the inversion process.

\paragraph{Improvement on Disentanglement.} As an exciting direction to overcome the limitation of the improvement of disentanglement, we plan to explore a way of constraining the latent space of NeRF-based GAN models to exhibit such disentanglement properties. Similar paths have been recently proposed for style-based generators~\cite{han2022ae}.

\paragraph{Finding Semantic Edits by Unsupervised or Self-supervised Manner.} This study is one of the pioneers for conducting semantic edits in 3D-aware generative models. Therefore, future studies can adapt the 2D unsupervised and self-supervised image manipulators like~\cite{yuksel2021latentclr,shen2021closed,patashnik2021styleclip}, to provide unsupervised methods for finding semantic edits.

\end{document}